\documentclass[pdflatex,sn-mathphys-num]{sn-jnl}

\usepackage{graphicx}%
\usepackage{multirow}%
\usepackage{amsmath,amssymb,amsfonts}%
\usepackage{amsthm}%
\usepackage{mathrsfs}%
\usepackage[title]{appendix}%
\usepackage{xcolor}%
\usepackage{textcomp}%
\usepackage{manyfoot}%
\usepackage{booktabs}%
\usepackage{algorithm}%
\usepackage{listings}%
\usepackage{algorithmic}
\usepackage{graphics}
\usepackage{anyfontsize}
\usepackage{physics}
\usepackage{bm}
\usepackage{hyperref}
\usepackage[normalem]{ulem}
\usepackage{makecell}
\usepackage{xcolor}

\useunder{\uline}{\ul}{}

\theoremstyle{thmstyleone}%
\theoremstyle{thmstyletwo}%

\theoremstyle{thmstylethree}%

\begin{document}

\title[SemICP]{SemICP: Semantic Non-Rigid Point Cloud Registration with Elastic Energy Regularization}

\author*[1,2]{\fnm{Wanwen} \sur{Chen}}\email{wanwenc@ece.ubc.ca}

\author[1]{\fnm{Qi} \sur{Zeng}}

\author[2]{\fnm{Carson} \sur{Studders}}

\author[3]{\fnm{Zongze} \sur{Li}}

\author[2]{\fnm{Jamie J.Y.} \sur{Kwon}}

\author[3]{\fnm{Tara} \sur{Kemper}}

\author[2]{\fnm{Emily H.T.} \sur{Pang}}

\author[2]{\fnm{Eitan} \sur{Prisman}}

\author[1,3]{\fnm{Septimiu E.} \sur{Salcudean}}

\affil[1]{\orgdiv{Department of Electrical and Computer Engineering}, \orgname{University of British Columbia}, \orgaddress{\city{Vancouver}, \state{BC}, \country{Canada}}}

\affil[2]{\orgdiv{Faculty of Medicine}, \orgname{University of British Columbia}, \orgaddress{\city{Vancouver}, \state{BC}, \country{Canada}}}

\affil[3]{\orgdiv{School of Biomedical Engineering}, \orgname{University of British Columbia}, \orgaddress{\city{Vancouver}, \state{BC}, \country{Canada}}}

\abstract{

\textbf{Purpose:} 
Accurate point cloud registration is essential in computer-aided interventions (CAI) to align multi-modal medical images for intraoperative guidance. 
Classical methods, such as Iterative Closest Point (ICP), remain attractive for their explainability and minimal training requirements, but typically ignore anatomical semantics and biomechanical properties during regularization.

\textbf{Methods:}
We present Semantic ICP (SemICP), a novel non-rigid point cloud registration framework that combines semantically informed point matching with linear elastic energy regularization. 
Semantic labels are used to improve correspondence matching by constraining correspondences to be anatomically consistent. 
A novel control-point deformation representation with linear-elastic energy regularization is introduced to encourage biomechanically plausible deformations. 
SemICP was evaluated on a finite element simulation-generated phantom and four datasets on US-CT, MR-CT, MR-MR and MR-US registration against established baselines. 
It was also tested with labels from AI-based segmentation in a fully automatic segmentation-registration pipeline.

\textbf{Results:}
On the finite element phantom, SemICP recovered 86\% of the interior displacement from surface observations alone.
On the real datasets, SemICP achieved the lowest target registration error on Liver MR-US
and the lowest surface distances on TORS, and ranked second on AbdominalMR-CT and HippocampusMR, where NR-ICP achieved lower surface distances. The fully automatic registration pipeline was shown to be effective for US-MR registration and to improve the alignment of expert-annotated structures.

\textbf{Conclusion:} 
SemICP improves interior accuracy at competitive surface accuracy by combining semantic correspondence constraints and linear energy regularization.
Combined with AI-based segmentation, SemICP provides an effective pipeline for multi-modal registration in CAI.
}


\maketitle

\section{Introduction}
\label{sec:introduction}
Registration is a fundamental challenge in CAI, where preoperative planning data must be aligned with intraoperative imaging.
In ultrasound (US)-guided navigation, registration remains challenging because US provides partial observations of internal structures.
Point clouds offer a modality-agnostic, sparse, and flexible anatomical structure representation, making them a common basis in US registration~\cite{ma2022augmented,hiep2025real}.

Learning-based registration shows promising performance, but struggles to generalize across domains~\cite{weber2024deep,zhang2024point}. 
Therefore, optimization-based registration, particularly ICP~\cite{besl1992method}, is widely used in CAI because of its simplicity and transferability. 
ICP variants have improved robustness~\cite{yang2015go,zhang2021fast} and can handle non-rigid registration~\cite{amberg2007optimal,yao2020quasi,yao2023fast}.

Despite their success, existing ICP-based methods have two key limitations. 

First, they establish correspondences primarily based on geometric proximity, ignoring semantic information such as anatomical labels.
Semantic constraints have improved rigid registration in autonomous driving~\cite{zaganidis2018integrating,wang2020robust} and concurrently in bone CT-X-ray rigid registration~\cite{flepp2025automatic}, but neither addresses surface deformation.
Registering each anatomy separately~\cite{hiep2025real} neglects tissue interactions. 
Consequently, semantic integration into non-rigid point cloud registration remains unexplored.

Second, existing non-rigid methods rely on local affine regularization, ignoring the underlying biomechanical effects. 
Direct finite element modeling (FEM) integrates biomechanical constraints~\cite{ringel2023regularized,yang2024boundary} but is computationally intensive. 
Data-driven priors that learn from simulation~\cite{fu2021deformable,wang2024libr+} require organ-specific training. 
Explicit elastic energy regularization, established in 3D image registration~\cite{nir2013model}, has not been extended to point clouds, which only provide boundary observations of the deformation field.

We propose Semantic ICP (SemICP), a non-rigid point cloud registration framework that combines semantic correspondence constraints with linear elastic energy regularization. SemICP exploits anatomical labels to guide correspondence estimation and introduces a control-point-based deformation representation, enabling efficient linear elastic energy regularization within an optimization-based registration framework.
We validate SemICP on a finite element phantom with known interior deformation and on four clinical datasets, demonstrating plausible volumetric deformation, improved interior alignment and compatibility with learning-based labelers.

\section{Methods}\label{sec:methods}
\begin{figure*}[tb]
\centering
    \includegraphics[trim={0 7.5cm 0 7.5cm},clip,width=\textwidth]{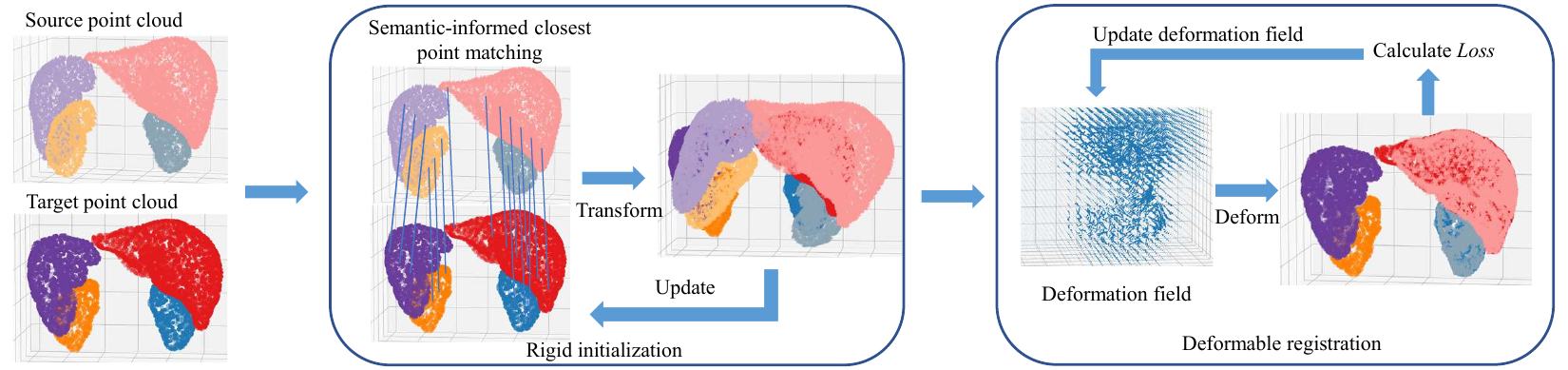}
    \caption{SemICP workflow: rigid initialization followed by non-rigid refinement.}\label{fig:workflow}
\end{figure*}

We define the source point cloud as $P = \{\bm{p_0},\bm{p_1},...,\bm{p_N}\}$ and the target as $Q = \{\bm{q_0},\bm{q_1},...,\bm{q_M}\}$, with $\bm{p},\bm{q} \in \mathbb{R}^3$, semantic labels $L_p=\{x_0,x_1,...,x_N\}$ and $L_q=\{y_0,y_1,...,y_M\}$ from label set $L=\{l_1,l_2...,l_K\}$, and local normals:  $N_p=\{\bm{n_0},...,\bm{n_N}\}$, $N_q=\{\bm{m_0},...,\bm{m_M}\}$.
$K$ is the number of labels, and we assume $K\geq2$.
Semantic-ICP (SemICP) first estimates a rigid transform $\bm{T} \in SE(3)$, then refines it by non-rigid registration (Fig.~\ref{fig:workflow}).

\subsection{Rigid Initialization}
The optimal rigid transform $\bm{T} \in SE(3)$ is estimated by minimizing a cost function that measures the point cloud difference (Algorithm~\ref{alg:rigid_init}).
At each iteration $i$, for each point $\bm{p_j}$ from $P$, we find the closest point $\bm{q_j}$ in $Q$ that has the smallest $L_2$ distance to points with the same semantic label.
KD-tree is used to accelerate the search for matched point pairs.
An Adam optimizer minimizes the cost function $L_{rigid}$ that measures the point-to-plane distance~\cite{segal2009generalized} to estimate  $\Delta \bm{T}^i$:
\begin{equation}\label{eq:rigid_loss}
L_{rigid} = \sum_{j=1}^N |(\bm{R}\bm{p_j}^{i}+\bm{t}-\bm{q_j}^{i})\cdot \bm{m_j}|^2
\end{equation}
where $\bm{t}$ is translation, $\bm{R}$ is rotation parametrized by XYZ Euler angles and $m_j$ is the surface normal of point $\bm{q_j}$.
The matched point pairs are recomputed at each iteration with the updated source point cloud until $L_{rigid}$ is no longer improved or the maximum number of iterations is reached.

 \begin{algorithm}
 \caption{Rigid initialization}\label{alg:rigid_init}
 \begin{algorithmic}[1]
 \renewcommand{\algorithmicrequire}{\textbf{Input:}}
 \renewcommand{\algorithmicensure}{\textbf{Output:}}
 \REQUIRE Source  $P$ and target  $Q$. 
 \ENSURE  Rigid transformation $\bm{T}$.
 \STATE Initialization: $\bm{T}^0=\bm{I}, P^0=P$.
  \FOR {$i = 0$ to $max\_iter-1$}
    \FOR {$l\in L$} 
        \STATE Subsets $P^i_{sub}$, $Q_{sub}$ with label $l$. 
        \FOR {$\bm{p_j}^i$ in $P^i_{sub}$}
            \STATE Find closest point $\bm{q_j}^i$ with smallest $L_2$ distance. 
        \ENDFOR
    \ENDFOR
  \STATE $\Delta \bm{T}^i \leftarrow \arg\min L_{rigid}(\bm{T}^i)$
  \STATE $\bm{T}^{i+1} \leftarrow \Delta \bm{T}^i\bm{T}^{i}$, $P^{i+1} \leftarrow \Delta \bm{T}^i P^{i}$
  \IF {Convergence Criteria Met}
    \STATE break
  \ENDIF
  \ENDFOR
 \RETURN $\bm{T}^{final}$
 \end{algorithmic}
 \end{algorithm}

\subsection{Non-rigid Registration}
Non-rigid registration estimates a displacement field that aligns $P$ with $Q$.
Instead of estimating a displacement for every surface point, we parameterize the
field by $N_C$ control points $\bm{C}=\{\bm{c_i}=(x_1^i,x_2^i,x_3^i),i=1,2,...,N_C\}$ placed on a regular
$n\times n\times n$ grid ($N_C=n^3$) with spacing $\Delta$. The unknowns
are the control-point displacements $\bm{D}\in\mathbb{R}^{N_C\times 3}$, whose
$i$-th row $\bm{d}_{\bm{c}_i}$ is the displacement of $\bm{c}_i$. They define a
continuous displacement field $\bm{D}(\bm{x})$ by trilinear interpolation over the
8 vertices of the grid cell containing $\bm{x}$.
The deformation is iteratively estimated per Algorithm~\ref{alg:non_rigid_refine} by minimizing:
\begin{equation}\label{eq:nr_loss}
L_{non-rigid}(\bm{D})= \frac{1}{N}\sum_{j=1}^N ||\bm{p_j}^{i}+\bm{d_j}^{i}-\bm{q_j}^{i}||_2^2 + \alpha Reg_{els}(\bm{D})
\end{equation}
where $\bm{d}_j$ is the deformation at the surface point $\bm{p}_j$, computed by trilinear interpolation over the 8 vertices of the grid cell containing $\bm{p}_j$.

The first term in Eq.~\eqref{eq:nr_loss} minimizes the matched point pair distance, while the remaining regularization term includes an elastic energy $Reg_{els}$.
$Reg_{els}$ is linear elastic energy regularization~\cite{nir2013model} that is approximated by two-point numerical differentiation of the Navier-Lam\'e Equation with step size $\Delta$:
\begin{equation}
Reg_{els}(\bm{D})=\frac{\Delta}{N_C}\sum_{\bm{c}}^{\bm{C}}(\frac{\mu}{4}\sum_{j=1}^3\sum_{k=1}^3 (\frac{\partial \bm{D}_j}{\partial x_k}+\frac{\partial \bm{D}_k}{\partial x_j})^{2} + \frac{\lambda}{2}(\nabla \cdot\bm{D})^2 )
\end{equation}
where $\lambda$ and $\mu$ are the Lam\'e parameters derived from Young's modulus $E$ and Poisson's ratio $\nu$:
\begin{equation}\label{eq:lame-parameter}
\lambda = \frac{E\nu}{(1+\nu)(1-2\nu)} \,\,,\,\, \mu = \frac{E}{2(1+\nu)}
\end{equation}

We use a two-point estimation for numerical differentiation:
\begin{equation}
\left.\frac{\partial D_j}{\partial x_k}\right|_{\bm{c}} = \frac{d_{\bm{c}+\Delta\bm{e}_k,\,j}-d_{\bm{c},\,j}}{\Delta},
\end{equation}
where $\bm{e}_k$ is the unit vector along axis k, and the difference is set to zero on the last grid layer.

 \begin{algorithm}
 \caption{Non-rigid refinement}\label{alg:non_rigid_refine}
 \begin{algorithmic}[1]
 \renewcommand{\algorithmicrequire}{\textbf{Input:}}
 \renewcommand{\algorithmicensure}{\textbf{Output:}}
 \REQUIRE Source  $P$ and target  $Q$. 
 \ENSURE  Deformation field $\bm{D}$.
 \STATE Initialization: $\bm{D}^0=\bm{0}$.
  \FOR {$i = 0$ to $max\_iter-1$}
    \FOR {$j \in N$}
        \STATE $\bm{d_j}^i = interp(\bm{D}^i)\ at\ \bm{p_j}$,  $\bm{p_j}^i=\bm{p_j} + \bm{d_j}^i$
    \ENDFOR
  \STATE  Find point pairs $(\bm{p_j}^i,\bm{q_j}^i)$ (lines 3 to 8 in Algorithm~\ref{alg:rigid_init})
  \STATE $\Delta \bm{D}^i \leftarrow \arg\min L_{non-rigid}(\bm{D}^i)$
  \STATE $\bm{D}^{i+1} \leftarrow \Delta \bm{D}^i + \bm{D}^{i}$
  \IF {Convergence Criteria Met}
    \STATE break
  \ENDIF
  \ENDFOR
 \RETURN $\bm{D}^{final}$
 \end{algorithmic}
 \end{algorithm}

\section{Experiments and Results}~\label{sec:validation}
\subsection{Datasets} 
Experiments are performed on two private datasets (TORS, Liver MR-3DUS), and two public datasets from the Learn2Reg challenges~\cite{hering2022learn2reg}. The datasets are as follows:

\textbf{TORS:} This dataset contains 24 pairs of preoperative CT and intraoperative 3D freehand US of the neck from patients who underwent Trans-Oral Robot-assisted Surgery (TORS) at the Vancouver General Hospital (Vancouver, BC, Canada), using the acquisition protocol in~\cite{chen2023towards}.
This study received ethics approval from the University of British Columbia (UBC) Clinical Research Ethics Board (H19-04025). 
A medical student and a research assistant segmented the carotid artery, jugular vein, and larynx/mandibular bone in US and CT images and the segmented structures were reconstructed to 3D point clouds.

\textbf{AbdominalMRCT}~\cite{clark2013cancer}: The dataset contains 8 pairs of intra-patient abdominal magnetic resonance (MR) and CT images, with segmentation of liver, spleen, right kidney and left kidney.

\textbf{HippocampusMR}~\cite{simpson2019large}:
This dataset includes MR images from healthy and pathological subjects with hippocampal head and tail segmentation. We use 60 inter-patient registration pairs from the official validation splits.

\textbf{LiverMR-3DUS}~\cite{zeng2020TMI, zeng2024validation}: The private dataset contains 174 pairs of intra-subject MR and 3D US image data from 19 healthy subjects and 20 patients with chronic liver diseases. 
The study was approved by the UBC Clinical Research Ethics Board (H14-01964).
Gallbladder, inferior vena cava (IVC), right kidney, and major hepatic vessel branches were manually segmented by experts as registration surrogates.

\subsection{Compared Baselines}

We compare SemICP against rigid methods (ICP~\cite{besl1992method}, GO-ICP~\cite{yang2015go}, Fast-Robust-ICP (FR-ICP)~\cite{zhang2021fast}), non-rigid methods (NR-ICP~\cite{amberg2007optimal}, Fast-RNRR~\cite{yao2020quasi}, AMM-NRR~\cite{yao2023fast}), probabilistic registration (CPD~\cite{gatti2022pycpd}), off-the-shelf learning-based models Deep Closest Point (DCP)~\cite{wang2019deep}), and where image-based registration is applicable, KeyMorph~\cite{evan2022keymorph} and TransMorph~\cite{chen2022transmorph}.
For each optimization-based method, we used cross-validation to optimize the parameters. We use hold-one-patient-out cross-validation for TORS and MRCT, hold-one-pair-out for HippocampusMR, and three-patient-level-fold cross-validation for liver. The search grid, details on the cross-validation definition are reported in the supplementary Section S2, and implementation details are reported in the supplementary Section S3.

Registration accuracy is evaluated using the bidirectional 95th percentile Hausdorff distance (HD95) and mean surface distance (MSD).
Target registration error (TRE) is reported when paired landmarks are available in the LiverMR-3DUS. 
Deformation smoothness is assessed by the standard deviation of the logarithmic Jacobian determinant (SDLogJ).
For transformation $\varphi(\bm{x})=\bm{x}+\bm{D}(\bm{x})$, the determinant of the Jacobian matrix
$\bm{J}=\bm{I}+\nabla\bm{D}$ defines the local volume ratio: values above~1
indicate expansion, values below~1 compression, and $\det\bm{J}\le 0$ folding. A low SDLogJ indicates a spatially uniform local volume change.
We use the SDLogJ implementation in the Learn2Reg challenge~\cite{hering2022learn2reg}, which adds an offset of 3 to $\det\bm{J}$  and clips it before taking the logarithm. 
We compute it on the control grid with the official Learn2Reg evaluation code.
\footnote{\url{https://github.com/MDL-UzL/L2R}}

\subsection{Evaluation on Simulation}\label{sec:exp-phantom}
To assess deformation recovery, we use finite element analysis to simulate a homogeneous soft-tissue $60\times60\times60$~mm cube, containing a sphere and an ellipsoid, so the point cloud has three labeled surfaces. We simulated the deformation using quadratic tetrahedral elements and linear elasticity with the bottom face clamped. We set $\nu=0.49, E=1kPa$ and the finite element solution is mesh-converged to 0.02 mm on average.
The simulation was implemented using scikit-fem~\cite{skfem2020}.
We simulated six different conditions: (1) loads on the top face to cause 1.5 mm, 3 mm and 6 mm downward displacement of the top face (compression), and (2) loads on the top face that cause 1.5 mm, 3 mm and 6 mm lateral displacement of the top face (shear).  
We then sampled 30{,}000 points distributed in proportion to surface area to generate the source point cloud and the target point cloud for SemICP.
The source and target point clouds were sampled independently on the phantom surface before and after deformation, so no point correspondences exist. 
Each configuration was run with three independent samplings, resulting in 18 experiments.
The finite element displacement on a regular 2 mm lattice inside the cube (27{,}000 points) is used to evaluate the interior displacement estimation accuracy and is never seen by the registration. 
We report the mean displacement error (L2 distance) over this lattice as the interior error and the L2 distance of the deformed surface points from their true displacement as the surface error, to evaluate both interior deformation estimation accuracy and surface matching accuracy. 
To evaluate the local difference in volume changes, we also report the mean absolute error of the Jacobian determinant $|\Delta\det J|$. 

\begin{figure*}[t]
\centering
\begin{minipage}{0.48\textwidth}
  \centering
  \textbf{(a)}\\[2pt]
  \includegraphics[width=\linewidth]{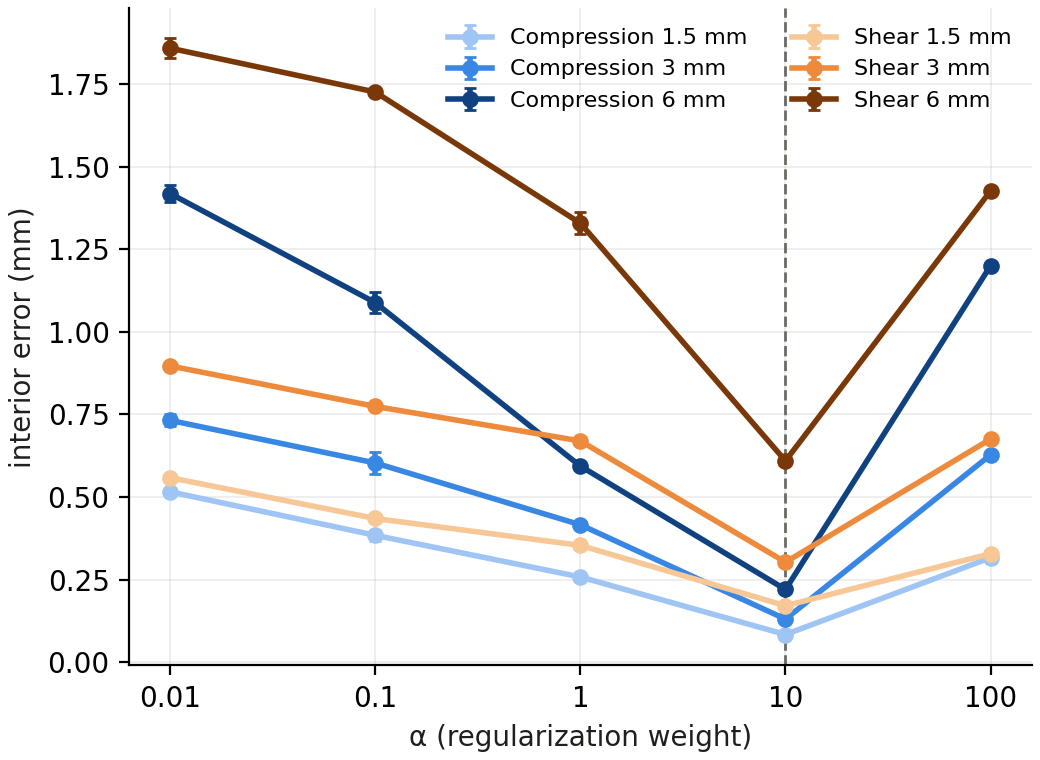}
\end{minipage}\\[6pt]
\begin{minipage}{\textwidth}
  \centering
  \textbf{(b)}\\[2pt]
  \includegraphics[width=\linewidth]{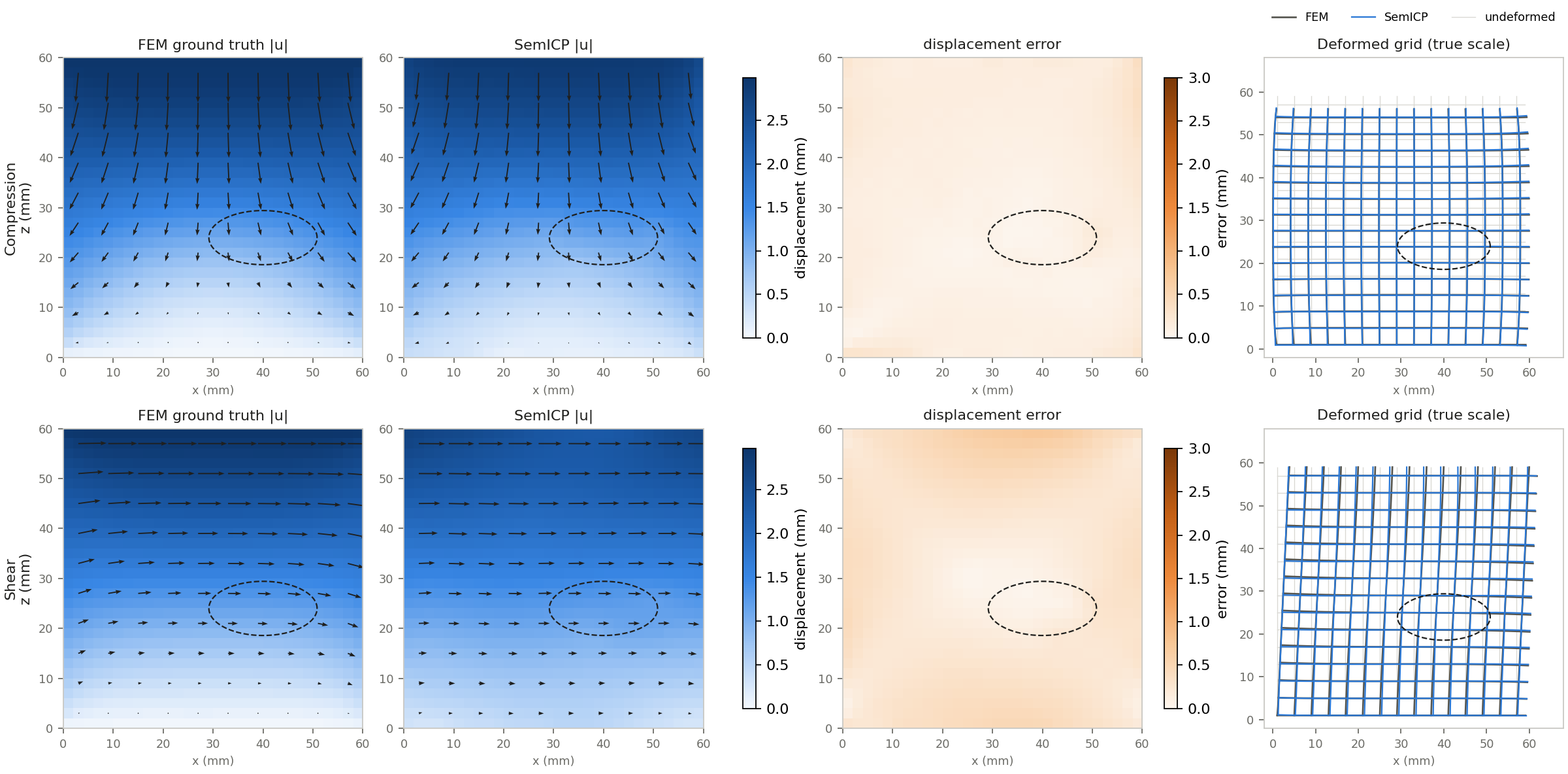}
\end{minipage}
\caption{SemICP on the finite element phantom. (a) Mean interior displacement error as a function of the regularization weight $\alpha$ for compression (blue) and shear (orange) of 1.5, 3 and 6\,mm (light to dark); mean $\pm$ standard deviation over three samplings; dashed: $\alpha=10$. 
(b) Recovered displacement at $\alpha=10$ for 3\,mm compression (top) and 3\,mm shear (bottom) on a vertical slice ($y=35$\,mm). Left to right: displacement magnitude of the finite element ground truth and of SemICP (arrows: in-plane displacement at twice its length), error magnitude, and deformed grid at true scale. Dashed: embedded structures.}
\label{fig:phantom}
\end{figure*}

We first evaluate the strength of the regularization weight $\alpha$. Fig.~\ref{fig:phantom} (a) shows the interior error as a function of $\alpha$, showing that $\alpha=10$ gives the lowest interior error for every load and magnitude. We therefore fix $\alpha=10$ for all the following experiments.

To separate the effect of the regularizer from that of the correspondence search, we also fit the same deformation grid with SemICP to the \emph{true} surface correspondences (\emph{oracle}).
Table~\ref{tab:phantom} shows that SemICP reduces the interior error from 1.85 to 0.25 mm on average for all loads and magnitudes (86\% of the displacement recovered). The error grows with the applied displacement but remains a nearly constant fraction of the applied displacement (7--10\% for compression, and about 20\% for shear). No folding occurred in any of the 18 phantom registrations (supplementary Table~S1). Under shear, the closest point matching has an inherent matching ambiguity since the top face moves tangentially to itself. 
With true correspondences (oracle in Table~\ref{tab:phantom}), SemICP recovers the interior displacement well (error fraction 4.0--4.3\% for compression and 1.5--1.8\% for shear).  Thus, the error in the shear loads largely comes from the correspondence search instead of the deformation model. Fig.~\ref{fig:phantom} (b) shows that under 3 mm compression and 3 mm shear, the recovered interior displacement error is small and the resulting deformation grids are close to FEM.  The results show that our SemICP can realistically recover the interior deformation.

\begin{table*}
\caption{Results on the FEM phantom. The mean interior and surface displacement errors are reported in mm, with the error as a percentage of the mean true displacement of the same points in parentheses. Oracle: SemICP with true surface correspondences in point matching. The surface error and Jacobian determinant error are reported on results estimated by SemICP. }\label{tab:phantom}
\centering
\resizebox{\textwidth}{!}{
\begin{tabular}{ll|c|cc|cc}
\hline
 & & Mean & \multicolumn{2}{c|}{Interior error, mm (\%)} & \multicolumn{2}{c}{SemICP} \\
Load & Mag. & displ.\ (mm) & SemICP & Oracle & Surface, mm (\%) & $|\Delta\det J|$ \\
\hline
Compr. & 1.5\,mm & 0.83 & 0.084 (10.1) & 0.034 (4.1) & 0.138 (16.5) & 0.003  \\
       & 3\,mm   & 1.66 & 0.131 (7.9)  & 0.071 (4.3) & 0.212 (12.7) & 0.004  \\
       & 6\,mm   & 3.32 & 0.220 (6.6)  & 0.133 (4.0) & 0.355 (10.6) & 0.007  \\
Shear  & 1.5\,mm & 0.76 & 0.171 (22.6) & 0.013 (1.8) & 0.224 (29.4) & 0.002 \\
       & 3\,mm   & 1.51 & 0.303 (20.0) & 0.026 (1.7) & 0.384 (25.2) & 0.002  \\
       & 6\,mm   & 3.02 & 0.610 (20.2) & 0.045 (1.5) & 0.768 (25.2) & 0.003 \\
\hline
\multicolumn{2}{l}{All (18)} & 1.85 & 0.253 (14.6) & 0.054 (2.9) & 0.347 (19.9) & 0.004 \\
\hline
\end{tabular}
}
\end{table*}

\subsection{Evaluation on Datasets}
Quantitative evaluations are reported in Tables~\ref{tab:TORS-full-label-hd95} to~\ref{tab:liver-gt}.
SemICP achieves the lowest HD95 and MSD on TORS, the second-lowest HD95 and MSD on AbdominalMRCT and HippocampusMR (after NR-ICP), and the lowest TRE and HD95 on Liver MR-US with an MSD comparable to NR-ICP.

\begin{figure*}
    \centering
    \includegraphics[width=\textwidth]{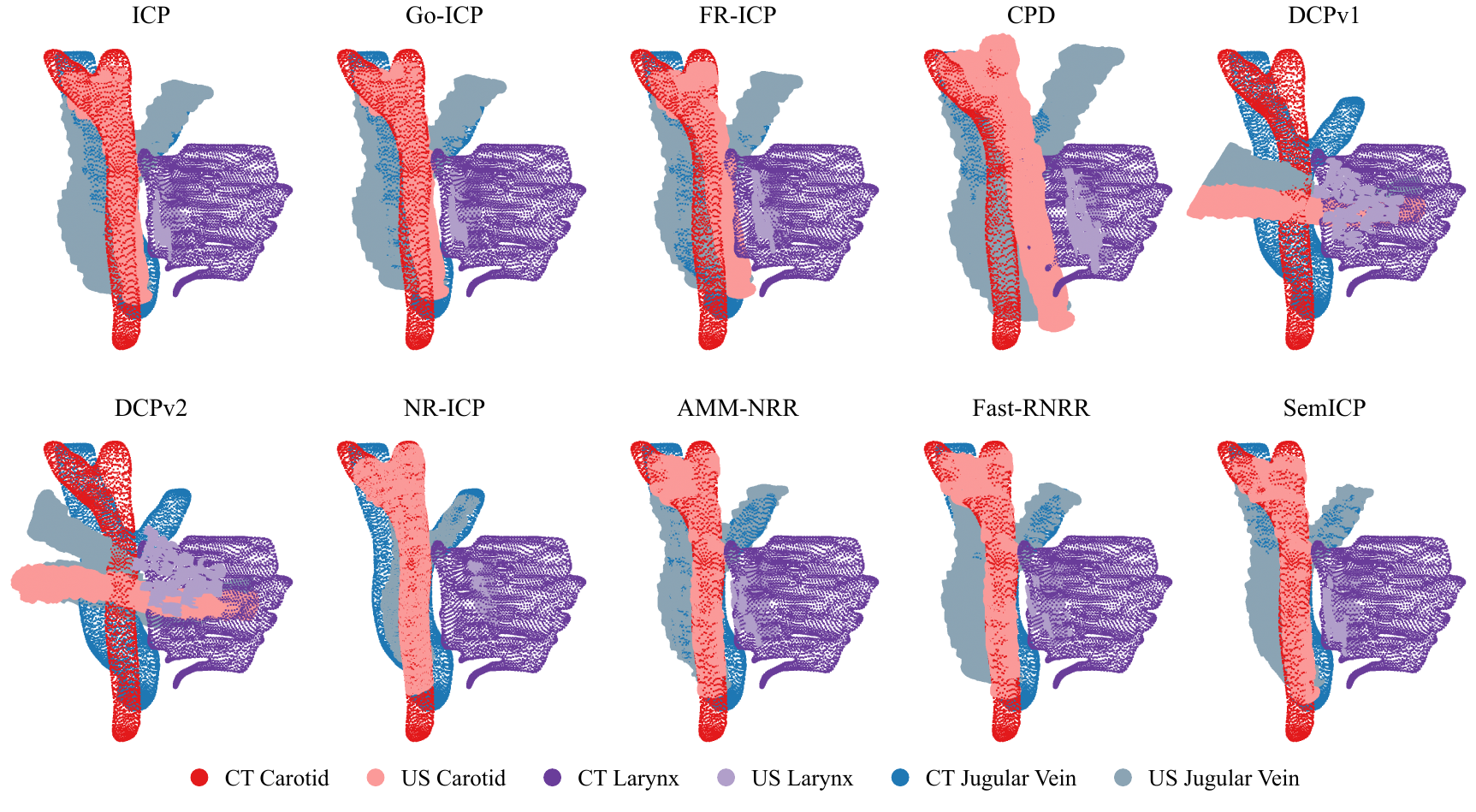}
    \caption{Registration results on a TORS US-CT pair (posterior-anterior view). Dark colors: CT structures (fixed). Light colors: US structures after registration by each method (moving). Each structure has one hue: red, carotid artery; purple, larynx/mandibular bone; blue, jugular vein.}\label{fig:TORS-real-example}
\end{figure*}

\textbf{TORS (Table~\ref{tab:TORS-full-label-hd95})}:
This dataset has partial US visibility of the larynx and bony structures, resulting in incomplete and imbalanced US point clouds.
AMM-NRR and Fast-RNRR completed registration for only 7 and 6 of 24 pairs, respectively. The node-sampling initialization of AMM-NRR and Fast-RNRR did not finish within the 15-minute limit for 17 pairs, which may be because the meshes consist of many small disconnected parts with
near-degenerate triangles. One Fast-RNRR result and one NR-ICP result diverged.
Methods relying solely on geometric proximity produce mismatches between different structures, as demonstrated in Fig.~\ref{fig:TORS-real-example}.
Learning-based DCP variants degrade surface matching accuracy, suggesting limited out-of-domain generalization.
SemICP achieves the best surface matching accuracy with an average runtime of 66 seconds, supporting its feasibility in image-guided intervention.

\begin{table*}
    \centering
    \caption{HD95 and MSD in mean (std), mm and runtime (s) on the TORS dataset.}\label{tab:TORS-full-label-hd95}
\resizebox{\textwidth}{!}{

\begin{tabular}{lcccccccccccc}
\toprule
 & Initial & DCPv1 & DCPv2 & ICP & GO-ICP & CPD & FR-ICP & NR-ICP & AMM-NRR & Fast-RNRR &  SemICP \\
\midrule
HD95 & \makecell{17.5\\(12.8)} & \makecell{30.3\\(10.8)} & \makecell{30.3\\(11.3)} & \makecell{15.6\\(11.6)} & \makecell{15.7\\(12.3)} & \makecell{16.4\\(11.1)} & \makecell{16.9\\(11.5)} & \makecell{17.1\\(11.7)} & \makecell{14.0\\(11.6)} & \makecell{15.6\\(11.8)}  & \makecell{13.5\\(12.9)} \\
\midrule
MSD & \makecell{6.8\\(4.4)} & \makecell{14.8\\(7.2)} & \makecell{14.6\\(7.9)} & \makecell{5.4\\(3.9)} & \makecell{5.4\\(3.9)} & \makecell{5.7\\(3.6)} & \makecell{5.9\\(4.2)} & \makecell{6.0\\(4.0)} & \makecell{4.4\\(2.8)} & \makecell{5.6\\(3.7)} & \makecell{4.1\\(3.0)} \\
\midrule
Time &  & \makecell{0.1\\(0.4)} & \makecell{0.1\\(0.4)} & \makecell{0.3\\(0.3)} & \makecell{12.7\\(0.4)} & \makecell{3.5\\(0.9)} & \makecell{15.3\\(8.7)} & \makecell{262.2\\(373.5)} & \makecell{53.0\\(14.8)} & \makecell{14.2\\(7.4)} & \makecell{65.8\\(24.1)} \\
\bottomrule
\end{tabular}
}

\end{table*}

\begin{figure*}
    \centering
    \includegraphics[width=\textwidth]{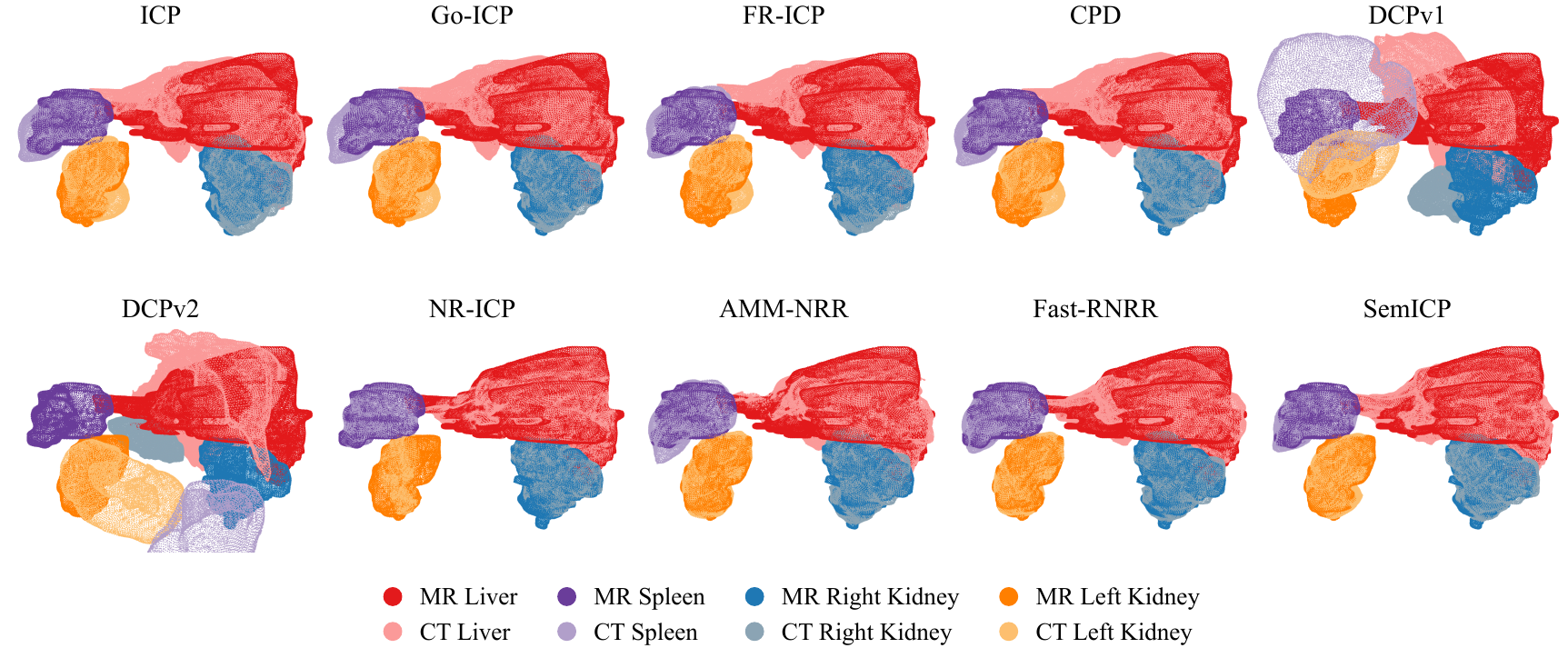}
    \caption{Registration results on one CT-MR pair in AbdominalMRCT (posterior-anterior view). Dark colors: MR structures (fixed). Light colors: CT structures after registration by each method (moving). Each structure has one hue: red, liver; purple, spleen; blue, right kidney; orange, left kidney.
}\label{fig:learn2regMRCT-real-example}
\end{figure*}

\textbf{AbdominalMRCT (Table~\ref{tab:learn2regMRCT-full-label-hd95}, Fig.~\ref{fig:learn2regMRCT-real-example})}: 
This dataset includes organs that undergo significant deformation.
Results for AMM-NRR and Fast-RNRR are reported for only 7 MR-CT pairs due to the initialization issues described above, while DCP variants again show poor accuracy due to domain mismatch.
SemICP achieved the second lowest HD95 and MSD with an average runtime of approximately 84 seconds, just after NR-ICP which needs 17.6 minutes, demonstrating its ability to handle large non-rigid deformations efficiently. 
SemICP trades surface accuracy for deformation regularity. In the regularization term ablation study (supplementary Table S2), without regularization, SemICP fits the surfaces more closely (HD95 7.4\,mm) but folds 7.9\% of the control grid.

\begin{table*}

     \caption{HD95 and MSD in mean (std), mm and runtime (s) on AbdominalMRCT. }\label{tab:learn2regMRCT-full-label-hd95}
    \resizebox{\textwidth}{!}{
\begin{tabular}{lccccccccccc}
\toprule
 & Initial & DCPv1 & DCPv2 & ICP & GO-ICP & CPD & FR-ICP & NR-ICP & AMM-NRR & Fast-RNRR & SemICP  \\
\midrule
HD95 & \makecell{38.4\\(19.4)} & \makecell{73.6\\(28.1)} & \makecell{96.0\\(54.8)} & \makecell{18.7\\(6.9)} & \makecell{18.9\\(7.5)} & \makecell{17.1\\(11.5)} & \makecell{23.7\\(10.4)} & \makecell{8.2\\(7.8)} & \makecell{18.4\\(17.6)} & \makecell{17.9\\(19.9)}  & \makecell{11.1\\(8.9)} \\
\midrule
MSD & \makecell{17.1\\(10.4)} & \makecell{37.1\\(22.5)} & \makecell{58.5\\(48.0)} & \makecell{7.2\\(2.5)} & \makecell{7.2\\(2.7)} & \makecell{7.3\\(5.8)} & \makecell{8.0\\(3.6)} & \makecell{2.2\\(1.6)} & \makecell{6.4\\(10.0)} & \makecell{6.8\\(9.5)} & \makecell{3.5\\(1.3)} \\
\midrule
Time &  & \makecell{0.8\\(1.9)} & \makecell{0.8\\(1.9)} & \makecell{1.1\\(0.5)} & \makecell{30.9\\(32.6)} & \makecell{25.4\\(38.4)} & \makecell{24.7\\(14.7)} & \makecell{1053.3\\(568.3)} & \makecell{91.6\\(24.7)} & \makecell{27.2\\(19.4)} & \makecell{84.1\\(12.2)} \\
\bottomrule
\end{tabular}
}
\end{table*}

\begin{figure}
\centering
\includegraphics[width=\textwidth]{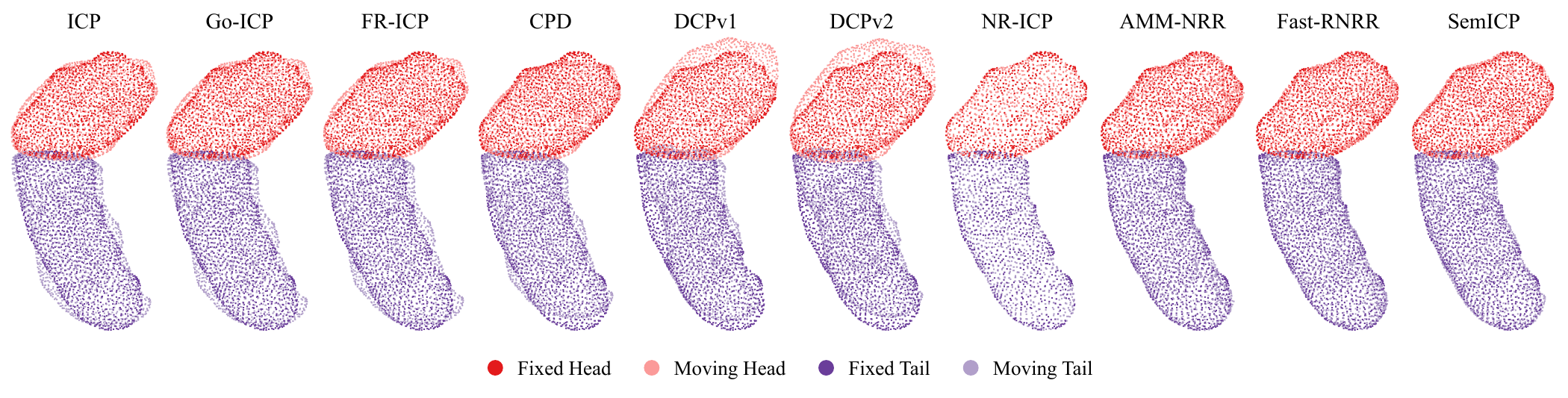}
\caption{ Registration results on one MR-MR pair in HippocampusMR (superior-inferior view). Dark colors: fixed MR structures. Light colors: moving MR structures after registration by each method. Each structure has one hue: red, hippocampus head; purple, hippocampus tail. }\label{fig:hippocampus-example}
\end{figure}

\textbf{HippocampusMR (Table~\ref{tab:learn2reg-hippocampus-full-label-hd95}, Fig.~\ref{fig:hippocampus-example})}: 
This dataset has the head and tail of the hippocampus, which are relatively small structures undergoing limited deformation compared to abdominal organs. 
SemICP yields the second-lowest HD95 and MSD after NR-ICP. AMM-NRR and Fast-RNRR completed registration on 58 and 57 of 60 pairs, respectively, with two pairs excluded due to the initialization issues described above, and one pair in Fast-RNRR crashed. 

\begin{table*}
    \caption{HD95 and MSD in mean (std), mm and runtime (s) on hippocampusMR. }\label{tab:learn2reg-hippocampus-full-label-hd95}
\resizebox{\textwidth}{!}{
\begin{tabular}{lcccccccccccc}
\toprule
 & Initial & DCPv1 & DCPv2 & ICP & GO-ICP & CPD & FR-ICP & NR-ICP & AMM-NRR & Fast-RNRR & SemICP \\
\midrule
HD95 & \makecell{3.8\\(1.3)} & \makecell{11.2\\(5.8)} & \makecell{13.5\\(5.7)} & \makecell{2.5\\(1.1)} & \makecell{2.6\\(1.3)} & \makecell{2.2\\(1.0)} & \makecell{2.8\\(1.4)} & \makecell{1.1\\(0.8)} & \makecell{2.2\\(1.7)} & \makecell{1.8\\(1.6)} & \makecell{1.1\\(1.2)} \\
\midrule
MSD & \makecell{1.7\\(0.6)} & \makecell{5.0\\(3.3)} & \makecell{6.3\\(3.5)} & \makecell{1.0\\(0.3)} & \makecell{1.0\\(0.3)} & \makecell{0.9\\(0.2)} & \makecell{1.0\\(0.3)} & \makecell{0.2\\(0.2)} & \makecell{0.6\\(0.3)} & \makecell{0.5\\(0.3)} & \makecell{0.4\\(0.2)} \\
\midrule
Time &  & \makecell{0.07\\(0.25)} & \makecell{0.08\\(0.26)} & \makecell{0.15\\(0.03)} & \makecell{25.2\\(17.9)} & \makecell{32.0\\(13.8)} & \makecell{0.16\\(0.06)} & \makecell{20.2\\(5.0)} & \makecell{0.7\\(0.2)} & \makecell{0.3\\(0.08)} & \makecell{13.8\\(1.4)} \\
\bottomrule
\end{tabular}
}

\end{table*}

\begin{figure}
\centering
\includegraphics[width=\textwidth]{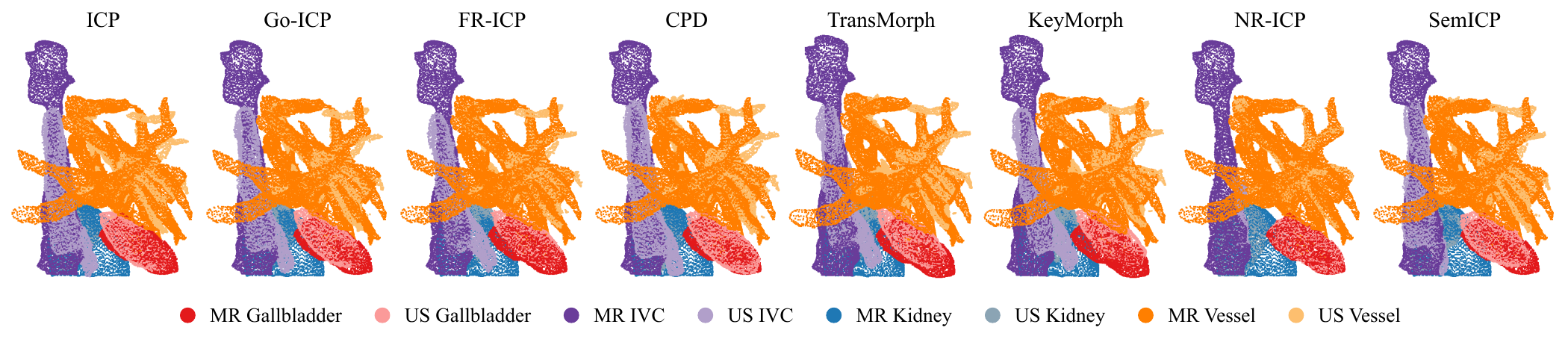}
\caption{Registration results on one MR-US pair in Liver MR-US (anterior-posterior view). Dark colors: MR structures (fixed). Light colors: US structures after registration by each method (moving). Each structure has one hue: red, gallbladder; purple, IVC; blue, kidney; orange, vessel. }\label{fig:liver-gt-example}
\end{figure}

\textbf{Liver MR-US (Table~\ref{tab:liver-gt}, Fig.~\ref{fig:liver-gt-example}):}
This dataset enables evaluation against KeyMorph and TransMorph trained from scratch via three-fold cross-validation. The dataset has 16 subjects and 10 different rigid initial misalignments, resulting in 160 test cases in total.
TRE at expert-annotated landmarks serves as an additional metric. 
SemICP achieves the lowest TRE and HD95, with an MSD comparable to NR-ICP,  providing improved surface and internal volumetric alignment.
Notably, SemICP's control-point deformation field enables volumetric interpolation, allowing TRE evaluation at internal landmarks. This is unavailable in NR-ICP, which estimates deformation only on surface points. Fast-RNRR and AMM-NRR were excluded because on eight randomly selected cases from eight subjects, their node-sampling initialization did not finish within the 15-minute limit.

\begin{table*}[]
    \caption{TRE, HD95 and MSD in mean (std), mm and runtime (s) on Liver.}\label{tab:liver-gt}
     \resizebox{\textwidth}{!}{
\begin{tabular}{lcccccccccc}
\toprule
 & Initial & KeyMorph & TransMorph & ICP & GO-ICP & CPD & FR-ICP & NR-ICP  & SemICP \\
\midrule
TRE & \makecell{14.5\\(5.4)} & \makecell{7.4\\(5.2)} & \makecell{8.0\\(5.9)} & \makecell{10.3\\(8.4)} & \makecell{7.8\\(8.3)} & \makecell{11.5\\(7.9)} & \makecell{7.0\\(6.1)} & \makecell{N/A\\(N/A)}  & \makecell{5.5\\(3.0)} \\
\midrule
HD95 & \makecell{25.2\\(5.2)} & \makecell{20.0\\(7.6)} & \makecell{19.7\\(7.5)} & \makecell{18.3\\(7.0)} & \makecell{19.2\\(8.9)} & \makecell{18.1\\(6.8)} & \makecell{17.9\\(5.6)} & \makecell{21.6\\(6.1)} & \makecell{15.6\\(5.6)} \\
\midrule
MSD & \makecell{9.8\\(3.1)} & \makecell{6.8\\(4.2)} & \makecell{6.8\\(4.8)} & \makecell{5.5\\(2.5)} & \makecell{6.5\\(5.9)} & \makecell{5.9\\(2.8)} & \makecell{5.4\\(2.7)} & \makecell{4.0\\(1.6)} & \makecell{4.0\\(2.1)} \\
\midrule
Time &  & \makecell{2.2\\(0.1)} & \makecell{0.04\\(0.01)} & \makecell{1.0\\(0.6)} & \makecell{39.5\\(161.9)} & \makecell{17.7\\(21.6)} & \makecell{15.0\\(6.0)} & \makecell{290.5\\(174.5)} & \makecell{18.0\\(5.9)} \\
\bottomrule
\end{tabular}
}

\end{table*}

\subsection{Integration with Learned Labelers}

\begin{figure}
\centering
\includegraphics[trim={2cm 5cm 1.2cm 5cm},clip,width=\textwidth]{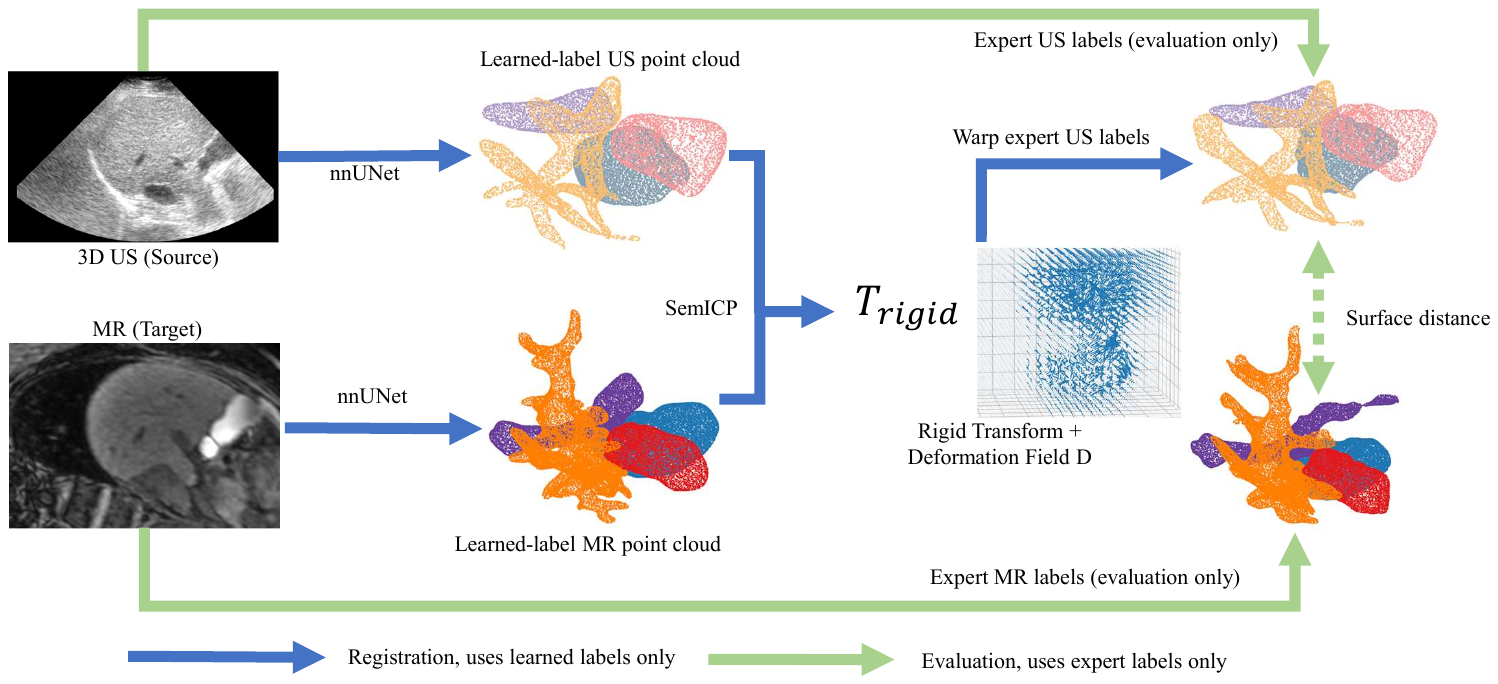}
\caption{The workflow overview of integration with learned labelers. }\label{fig:liver-nnunet-workflow}
\end{figure}

We evaluate SemICP with learned semantic labelers nnUNet~\cite{isensee2021nnu} to test whether SemICP still works when its semantic labels come from a segmentation network instead of an expert. The experiment was conducted on the Liver dataset via five-fold cross-validation. 
SemICP uses the same settings as in the other Liver experiments ($\alpha=10$, $E=1$, and $\nu=0.45$).
For each subject, the segmented US and MR volumes are converted to labeled point clouds, which serve as the source and target of SemICP, respectively. No expert labels are used during registration.
The estimated deformation is applied to warp the expert-labeled US point clouds, which are not part of the registration inputs. Because the control-point field is defined throughout the volume, it can warp these points even though they differ from the registered ones. The surface distance between the expert-labeled point clouds is reported as a measure of the proposed segmentation-registration workflow (Fig.~\ref{fig:liver-nnunet-workflow}).

\begin{figure}
\centering
\includegraphics[width=\textwidth]{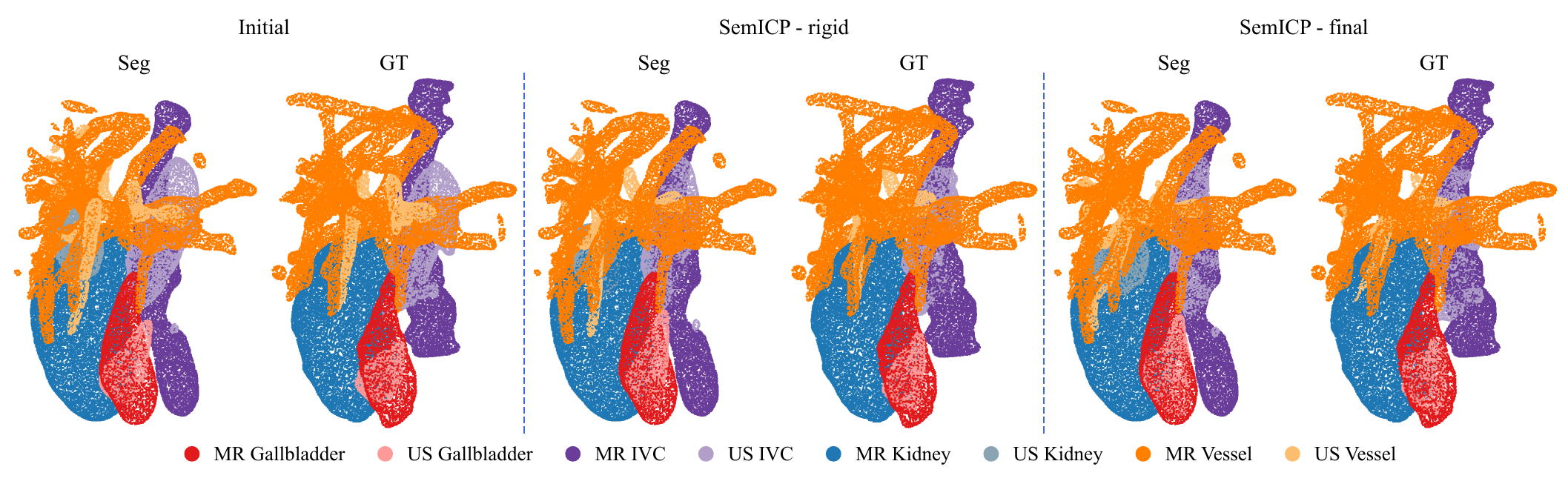}
\caption{Liver registration results with learned labelers (anterior-posterior view). Dark
colors: MR structures (fixed). Light colors: US structures after each registration stage (moving).
Each structure has one hue: red, gallbladder; purple, IVC; blue, kidney; orange, vessel.}\label{fig:liver-example}
\end{figure}

The nnUNet segmentation results in Table~\ref{tab:liver-segment} show higher Dice for MR than US due to lower US image quality. 
We report the unidirectional HD95 and MSD from the warped expert-labeled US points to the expert-labeled MR point cloud in Table~\ref{tab:liver-case-study} (row ``GT''), and for reference, the rows ``Seg'' report the residual between the registered learned point clouds to show how well SemICP fits its own input. An illustration is shown in Fig.~\ref{fig:liver-example}.
SemICP improves alignment of imperfect segmented point clouds, reducing the distance between the learned point clouds from 7.8--22.0\,mm to 3.7--4.5\,mm HD95, demonstrating robustness to label error. 
Applying the estimated deformation to expert-labeled point clouds enhances their alignment, reducing their HD95 from 9.4--15.6\,mm initially to 6.0--9.1\,mm, indicating that the pipeline improves MR–US registration under realistic segmentation uncertainty.

\begin{table*}[htbp]
\centering
\caption{nnUnet segmentation 5-fold evaluation results reported in mean (std). }\label{tab:liver-segment}
     \resizebox{\textwidth}{!}{
\begin{tabular}{lcccccc}
\toprule
\textbf{Dataset} & \textbf{Metrics} & \textbf{Gallbladder}& \textbf{IVC} & \textbf{Right Kidney} & \textbf{Vessels} & \textbf{Average} \\
\midrule
\multirow{2}{*}{\textbf{MR T1-W}} 
& Dice &  0.867(0.073) & 0.773(0.095) & 0.955(0.010) &  0.729(0.085) & 0.831(0.066)\\
& IoU  & 0.762(0.088) & 0.630(0.106) &  0.918(0.012)  & 0.578(0.096) & 0.722(0.075)\\
\midrule
\multirow{2}{*}{\textbf{US Bmode}} 
& Dice & 0.771(0.069) & 0.700(0.076) & 0.606(0.091) & 0.689(0.073) & 0.692(0.072)\\
& IoU  & 0.670(0.075) & 0.560(0.081) & 0.520(0.102) & 0.543(0.084) & 0.573(0.082)\\
\bottomrule
\end{tabular}
}
\label{tab:mr_us_seg_comparison}
\end{table*}

\begin{table*}[]
\centering
\caption{Unidirectional HD95 and MSD (mean$\pm$std, mm) on liver MR-3DUS. Seg: results on learning-based point clouds. GT: results on expert-labeled point clouds. }\label{tab:liver-case-study}
\resizebox{\textwidth}{!}{
\begin{tabular}{cc|cc|cc|cc}
\toprule
                              &     & \multicolumn{2}{c|}{Initial}       & \multicolumn{2}{c|}{SemICP-rigid}  & \multicolumn{2}{c}{SemICP-final}   \\
                              &     & HD95            & MSD             & HD95             & MSD            & HD95             & MSD             \\
                              \midrule
\multirow{2}{*}{Gallbladder}  & Seg & $11.21\pm12.15$ & $5.20\pm8.92$ & $8.92\pm7.21$ & $3.96\pm4.46$ & $3.67\pm1.29$ & $1.46\pm0.61$ \\
                              & GT  & $9.43\pm3.89$ & $4.19\pm1.90$ & $8.86\pm4.49$ & $3.99\pm2.68$ & $5.97\pm2.84$ & $2.51\pm1.32$ \\
                              \midrule
\multirow{2}{*}{IVC}          & Seg & $11.07\pm6.60$ & $4.54\pm2.73$ & $9.77\pm8.15$ & $4.16\pm3.63$ & $3.78\pm2.20$ & $1.44\pm0.42$ \\
                              & GT  & $11.39\pm7.42$ & $4.78\pm3.41$ & $11.48\pm10.52$ & $4.77\pm4.31$ & $8.23\pm6.93$ & $3.09\pm2.25$ \\
                              \midrule
\multirow{2}{*}{Kidney}       & Seg & $21.97\pm25.26$ & $12.14\pm24.16$ & $16.60\pm12.94$ & $7.38\pm11.86$ & $4.50\pm4.31$ & $1.91\pm4.03$ \\
                              & GT  & $15.63\pm5.17$ & $7.13\pm4.71$ & $13.99\pm3.23$ & $5.32\pm2.24$ & $7.82\pm4.36$ & $3.05\pm1.81$ \\
                              \midrule
\multirow{2}{*}{Liver vessel} & Seg & $7.82\pm5.00$ & $2.96\pm1.53$ & $9.16\pm6.98$ & $3.68\pm2.93$ & $3.72\pm1.17$ & $1.59\pm0.39$ \\
                              & GT  & $9.92\pm4.97$ & $3.52\pm1.58$ & $11.51\pm7.17$ & $4.37\pm3.09$ & $9.13\pm5.12$ & $3.02\pm1.50$ \\
                              \midrule
\multicolumn{2}{c|}{Time}            & \multicolumn{2}{c|}{}              & \multicolumn{2}{c|}{$11.44\pm1.79$} & \multicolumn{2}{c}{$36.27\pm7.76$} \\
\bottomrule
\end{tabular}
}
\end{table*}

\section{Discussion}~\label{sec:discussion}

The evaluated datasets present diverse challenges, including partial observations and class imbalance (TORS), large inter-modality displacement (AbdominalMRCT), small structures requiring precise alignment (HippocampusMR), and artificial rigid misalignment and complex vascular geometry (LiverMR--3DUS). 
All optimization-based baselines were tuned with the same cross-validation protocol as SemICP, and tuning mainly
improved NR-ICP and AMM-NRR, so the comparison does not favour SemICP through its parameter choice.
SemICP consistently achieves the lowest or second-lowest HD95 and MSD among all methods, and the lowest TRE on Liver MR-US, demonstrating the robustness of combining semantic correspondence constraints with linear elastic energy regularization.
NR-ICP achieves lower surface distances on AbdominalMRCT and HippocampusMR, but surface distance does not necessarily measure the interior accuracy. On the phantom, SemICP recovers the interior (Section~\ref{sec:exp-phantom}), and on Liver MR-US, SemICP has the lowest TRE although its MSD is not the lowest. Evaluating TRE together with HD95 and MSD gives
a more complete picture of registration performance.
Label-informed matching reduces mismatches that are common in geometry-based methods, as confirmed by the ablation study (supplementary Section S1.1), and has improved robustness as demonstrated by the sensitivity analysis in the supplementary Section S4. 

Compared with existing non-rigid ICP variants, our method models volumetric coherence and provides physically motivated elastic energy regularization, and the simulation study demonstrates that it recovers the interior displacement where the surfaces are observed. The ablation studies in the supplementary Section S1 further demonstrate its improved interior displacement accuracy and deformation plausibility compared with gradient-based and magnitude-based regularization.

DCP variants degrade surface alignment, confirming that models trained on everyday objects do not generalize to medical point clouds. SemICP outperforms learning methods KeyMorph and TransMorph, demonstrating that explicit anatomical and biomechanical priors can be competitive with learned approaches.

The LiverMR--3DUS experiment demonstrates that SemICP integrates effectively with learning-based segmentation and remains robust to imperfect segmentation. Crucially, the control-point deformation field enables interpolation of motion in unobserved regions, facilitating deformation transfer to internal tissues. It is a capability unavailable in surface-only non-rigid registration methods.

\textbf{Limitations:} SemICP requires semantic labels, and registration performance depends on segmentation quality.
This method is designed for multi-organ registration, but it may be generalized to single-organ registration with distinct intra-organ labels.
Elasticity parameters are currently homogeneous and may not be the actual value. Future work could assign tissue-specific biomechanical parameters per control point.
Closest-point correspondences cannot observe motion tangential to a surface, which limits the recovery of shear.

\section{Conclusion}~\label{sec:conclusion}
We present SemICP, a novel non-rigid point cloud registration framework combining semantic label-based correspondence with control point-based deformation representation and linear elastic energy regularization. SemICP recovers the interior deformation in FEM simulation studies from surface observations alone. SemICP achieves the lowest TRE on Liver MR-US and competitive surface accuracy on four diverse real-world datasets, and integrates effectively with AI segmentation, demonstrating robustness and potential in CAI.

\backmatter

\section*{Declarations}

\begin{itemize}
\item Funding: This work was supported by NSERC Discovery Grant and the Charles Laszlo Chair in Biomedical Engineering held by Dr. Salcudean, VCHRI Innovation and Translational Research Award, the University of British Columbia Department of Surgery Seed Grant, and CIHR Project Grant held by Dr. Prisman. 
\item Competing interests: The authors have no competing interests to declare that are relevant to the content of this article.
\item Ethics approval: Approved by the University of British Columbia Clinical Research Ethics Board (H19-04025, H14-01964). Informed consent was obtained from all participants. This study was performed in line with the principles of the Declaration of Helsinki.
\end{itemize}

\bibliography{sn-bibliography}% common bib file


\begin{thebibliography}{31}
\ifx \bisbn   \undefined \def \bisbn  #1{ISBN #1}\fi
\ifx \binits  \undefined \def \binits#1{#1}\fi
\ifx \bauthor  \undefined \def \bauthor#1{#1}\fi
\ifx \batitle  \undefined \def \batitle#1{#1}\fi
\ifx \bjtitle  \undefined \def \bjtitle#1{#1}\fi
\ifx \bvolume  \undefined \def \bvolume#1{\textbf{#1}}\fi
\ifx \byear  \undefined \def \byear#1{#1}\fi
\ifx \bissue  \undefined \def \bissue#1{#1}\fi
\ifx \bfpage  \undefined \def \bfpage#1{#1}\fi
\ifx \blpage  \undefined \def \blpage #1{#1}\fi
\ifx \burl  \undefined \def \burl#1{\textsf{#1}}\fi
\ifx \doiurl  \undefined \def \doiurl#1{\url{https://doi.org/#1}}\fi
\ifx \betal  \undefined \def \betal{\textit{et al.}}\fi
\ifx \binstitute  \undefined \def \binstitute#1{#1}\fi
\ifx \binstitutionaled  \undefined \def \binstitutionaled#1{#1}\fi
\ifx \bctitle  \undefined \def \bctitle#1{#1}\fi
\ifx \beditor  \undefined \def \beditor#1{#1}\fi
\ifx \bpublisher  \undefined \def \bpublisher#1{#1}\fi
\ifx \bbtitle  \undefined \def \bbtitle#1{#1}\fi
\ifx \bedition  \undefined \def \bedition#1{#1}\fi
\ifx \bseriesno  \undefined \def \bseriesno#1{#1}\fi
\ifx \blocation  \undefined \def \blocation#1{#1}\fi
\ifx \bsertitle  \undefined \def \bsertitle#1{#1}\fi
\ifx \bsnm \undefined \def \bsnm#1{#1}\fi
\ifx \bsuffix \undefined \def \bsuffix#1{#1}\fi
\ifx \bparticle \undefined \def \bparticle#1{#1}\fi
\ifx \barticle \undefined \def \barticle#1{#1}\fi
\bibcommenthead
\ifx \bconfdate \undefined \def \bconfdate #1{#1}\fi
\ifx \botherref \undefined \def \botherref #1{#1}\fi
\ifx \url \undefined \def \url#1{\textsf{#1}}\fi
\ifx \bchapter \undefined \def \bchapter#1{#1}\fi
\ifx \bbook \undefined \def \bbook#1{#1}\fi
\ifx \bcomment \undefined \def \bcomment#1{#1}\fi
\ifx \oauthor \undefined \def \oauthor#1{#1}\fi
\ifx \citeauthoryear \undefined \def \citeauthoryear#1{#1}\fi
\ifx \endbibitem  \undefined \def \endbibitem {}\fi
\ifx \bconflocation  \undefined \def \bconflocation#1{#1}\fi
\ifx \arxivurl  \undefined \def \arxivurl#1{\textsf{#1}}\fi
\csname PreBibitemsHook\endcsname

%%% 1
\bibitem[\protect\citeauthoryear{Ma et~al.}{2022}]{ma2022augmented}
\begin{barticle}
\bauthor{\bsnm{Ma}, \binits{L.}},
\bauthor{\bsnm{Liang}, \binits{H.}},
\bauthor{\bsnm{Han}, \binits{B.}},
\bauthor{\bsnm{Yang}, \binits{S.}},
\bauthor{\bsnm{Zhang}, \binits{X.}},
\bauthor{\bsnm{Liao}, \binits{H.}}:
\batitle{Augmented reality navigation with ultrasound-assisted point cloud registration for percutaneous ablation of liver tumors}.
\bjtitle{International journal of computer assisted radiology and surgery}
\bvolume{17}(\bissue{9}),
\bfpage{1543}--\blpage{1552}
(\byear{2022})
\end{barticle}
\endbibitem

%%% 2
\bibitem[\protect\citeauthoryear{Hiep et~al.}{2025}]{hiep2025real}
\begin{barticle}
\bauthor{\bsnm{Hiep}, \binits{M.}},
\bauthor{\bsnm{Heerink}, \binits{W.}},
\bauthor{\bsnm{Groen}, \binits{H.}},
\bauthor{\bsnm{Saiz}, \binits{L.A.}},
\bauthor{\bsnm{Grotenhuis}, \binits{B.}},
\bauthor{\bsnm{Beets}, \binits{G.}},
\bauthor{\bsnm{Aalbers}, \binits{A.}},
\bauthor{\bsnm{Kuhlmann}, \binits{K.F.}},
\bauthor{\bsnm{Ruers}, \binits{T.}}:
\batitle{Real-time intraoperative ultrasound registration for accurate surgical navigation in patients with pelvic malignancies}.
\bjtitle{International journal of computer assisted radiology and surgery}
\bvolume{20}(\bissue{2}),
\bfpage{249}--\blpage{258}
(\byear{2025})
\end{barticle}
\endbibitem

%%% 3
\bibitem[\protect\citeauthoryear{Weber et~al.}{2024}]{weber2024deep}
\begin{bchapter}
\bauthor{\bsnm{Weber}, \binits{M.}},
\bauthor{\bsnm{Wild}, \binits{D.}},
\bauthor{\bsnm{Kleesiek}, \binits{J.}},
\bauthor{\bsnm{Egger}, \binits{J.}},
\bauthor{\bsnm{Gsaxner}, \binits{C.}}:
\bctitle{Deep learning-based point cloud registration for augmented reality-guided surgery}.
In: \bbtitle{2024 IEEE International Symposium on Biomedical Imaging (ISBI)},
pp. \bfpage{1}--\blpage{5}
(\byear{2024}).
\bcomment{IEEE}
\end{bchapter}
\endbibitem

%%% 4
\bibitem[\protect\citeauthoryear{Zhang et~al.}{2024}]{zhang2024point}
\begin{barticle}
\bauthor{\bsnm{Zhang}, \binits{Y.}},
\bauthor{\bsnm{Zou}, \binits{Y.}},
\bauthor{\bsnm{Liu}, \binits{P.X.}}:
\batitle{Point cloud registration in laparoscopic liver surgery using keypoint correspondence registration network}.
\bjtitle{IEEE Transactions on Medical Imaging}
\bvolume{44}(\bissue{2}),
\bfpage{749}--\blpage{760}
(\byear{2024})
\end{barticle}
\endbibitem

%%% 5
\bibitem[\protect\citeauthoryear{Besl and McKay}{1992}]{besl1992method}
\begin{bchapter}
\bauthor{\bsnm{Besl}, \binits{P.J.}},
\bauthor{\bsnm{McKay}, \binits{N.D.}}:
\bctitle{Method for registration of 3-d shapes}.
In: \bbtitle{Sensor Fusion IV: Control Paradigms and Data Structures},
vol. \bseriesno{1611},
pp. \bfpage{586}--\blpage{606}
(\byear{1992}).
\bcomment{SPIE}
\end{bchapter}
\endbibitem

%%% 6
\bibitem[\protect\citeauthoryear{Yang et~al.}{2015}]{yang2015go}
\begin{barticle}
\bauthor{\bsnm{Yang}, \binits{J.}},
\bauthor{\bsnm{Li}, \binits{H.}},
\bauthor{\bsnm{Campbell}, \binits{D.}},
\bauthor{\bsnm{Jia}, \binits{Y.}}:
\batitle{Go-icp: A globally optimal solution to 3d icp point-set registration}.
\bjtitle{IEEE transactions on pattern analysis and machine intelligence}
\bvolume{38}(\bissue{11}),
\bfpage{2241}--\blpage{2254}
(\byear{2015})
\end{barticle}
\endbibitem

%%% 7
\bibitem[\protect\citeauthoryear{Zhang et~al.}{2022}]{zhang2021fast}
\begin{barticle}
\bauthor{\bsnm{Zhang}, \binits{J.}},
\bauthor{\bsnm{Yao}, \binits{Y.}},
\bauthor{\bsnm{Deng}, \binits{B.}}:
\batitle{Fast and robust iterative closest point}.
\bjtitle{IEEE Transactions on Pattern Analysis and Machine Intelligence}
\bvolume{44}(\bissue{7}),
\bfpage{3450}--\blpage{3466}
(\byear{2022})
\end{barticle}
\endbibitem

%%% 8
\bibitem[\protect\citeauthoryear{Amberg et~al.}{2007}]{amberg2007optimal}
\begin{bchapter}
\bauthor{\bsnm{Amberg}, \binits{B.}},
\bauthor{\bsnm{Romdhani}, \binits{S.}},
\bauthor{\bsnm{Vetter}, \binits{T.}}:
\bctitle{Optimal step nonrigid icp algorithms for surface registration}.
In: \bbtitle{2007 IEEE Conference on Computer Vision and Pattern Recognition (CVPR)},
pp. \bfpage{1}--\blpage{8}
(\byear{2007}).
\bcomment{IEEE}
\end{bchapter}
\endbibitem

%%% 9
\bibitem[\protect\citeauthoryear{Yao et~al.}{2020}]{yao2020quasi}
\begin{bchapter}
\bauthor{\bsnm{Yao}, \binits{Y.}},
\bauthor{\bsnm{Deng}, \binits{B.}},
\bauthor{\bsnm{Xu}, \binits{W.}},
\bauthor{\bsnm{Zhang}, \binits{J.}}:
\bctitle{Quasi-newton solver for robust non-rigid registration}.
In: \bbtitle{2020 IEEE/CVF Conference on Computer Vision and Pattern Recognition (CVPR)},
pp. \bfpage{7597}--\blpage{7606}
(\byear{2020})
\end{bchapter}
\endbibitem

%%% 10
\bibitem[\protect\citeauthoryear{Yao et~al.}{2023}]{yao2023fast}
\begin{barticle}
\bauthor{\bsnm{Yao}, \binits{Y.}},
\bauthor{\bsnm{Deng}, \binits{B.}},
\bauthor{\bsnm{Xu}, \binits{W.}},
\bauthor{\bsnm{Zhang}, \binits{J.}}:
\batitle{Fast and robust non-rigid registration using accelerated majorization-minimization}.
\bjtitle{IEEE Transactions on Pattern Analysis and Machine Intelligence}
\bvolume{45}(\bissue{8}),
\bfpage{9681}--\blpage{9698}
(\byear{2023})
\end{barticle}
\endbibitem

%%% 11
\bibitem[\protect\citeauthoryear{Zaganidis et~al.}{2018}]{zaganidis2018integrating}
\begin{barticle}
\bauthor{\bsnm{Zaganidis}, \binits{A.}},
\bauthor{\bsnm{Sun}, \binits{L.}},
\bauthor{\bsnm{Duckett}, \binits{T.}},
\bauthor{\bsnm{Cielniak}, \binits{G.}}:
\batitle{Integrating deep semantic segmentation into 3-d point cloud registration}.
\bjtitle{IEEE Robotics and automation letters}
\bvolume{3}(\bissue{4}),
\bfpage{2942}--\blpage{2949}
(\byear{2018})
\end{barticle}
\endbibitem

%%% 12
\bibitem[\protect\citeauthoryear{Wang et~al.}{2020}]{wang2020robust}
\begin{bchapter}
\bauthor{\bsnm{Wang}, \binits{Q.}},
\bauthor{\bsnm{Yang}, \binits{Y.}},
\bauthor{\bsnm{Wan}, \binits{T.}},
\bauthor{\bsnm{Du}, \binits{S.}}:
\bctitle{Robust point set registration based on semantic information}.
In: \bbtitle{2020 IEEE International Conference on Systems, Man, and Cybernetics (SMC)}
(\byear{2020}).
\bcomment{IEEE}
\end{bchapter}
\endbibitem

%%% 13
\bibitem[\protect\citeauthoryear{Flepp et~al.}{2025}]{flepp2025automatic}
\begin{botherref}
\oauthor{\bsnm{Flepp}, \binits{R.}},
\oauthor{\bsnm{Nissen}, \binits{L.}},
\oauthor{\bsnm{Sigrist}, \binits{B.}},
\oauthor{\bsnm{Nieuwland}, \binits{A.}},
\oauthor{\bsnm{Cavalcanti}, \binits{N.}},
\oauthor{\bsnm{F{\"u}rnstahl}, \binits{P.}},
\oauthor{\bsnm{Dreher}, \binits{T.}},
\oauthor{\bsnm{Calvet}, \binits{L.}}:
Automatic multi-view x-ray/ct registration using bone substructure contours.
International Journal of Computer Assisted Radiology and Surgery,
1--8
(2025)
\end{botherref}
\endbibitem

%%% 14
\bibitem[\protect\citeauthoryear{Ringel et~al.}{2023}]{ringel2023regularized}
\begin{bchapter}
\bauthor{\bsnm{Ringel}, \binits{M.}},
\bauthor{\bsnm{Heiselman}, \binits{J.}},
\bauthor{\bsnm{Richey}, \binits{W.}},
\bauthor{\bsnm{Meszoely}, \binits{I.}},
\bauthor{\bsnm{Miga}, \binits{M.}}:
\bctitle{Regularized kelvinlet functions to model linear elasticity for image-to-physical registration of the breast}.
In: \bbtitle{International Conference on Medical Image Computing and Computer-Assisted Intervention (MICCAI)},
pp. \bfpage{344}--\blpage{353}
(\byear{2023}).
\bcomment{Springer}
\end{bchapter}
\endbibitem

%%% 15
\bibitem[\protect\citeauthoryear{Yang et~al.}{2025}]{yang2024boundary}
\begin{barticle}
\bauthor{\bsnm{Yang}, \binits{Z.}},
\bauthor{\bsnm{Simon}, \binits{R.}},
\bauthor{\bsnm{Merrell}, \binits{K.}},
\bauthor{\bsnm{Linte}, \binits{C.A.}}:
\batitle{Boundary constraint-free biomechanical model-based surface matching for intraoperative liver deformation correction}.
\bjtitle{IEEE Transactions on Medical Imaging}
\bvolume{44}(\bissue{4}),
\bfpage{1723}--\blpage{1734}
(\byear{2025})
\end{barticle}
\endbibitem

%%% 16
\bibitem[\protect\citeauthoryear{Fu et~al.}{2021}]{fu2021deformable}
\begin{barticle}
\bauthor{\bsnm{Fu}, \binits{Y.}},
\bauthor{\bsnm{Wang}, \binits{T.}},
\bauthor{\bsnm{Lei}, \binits{Y.}},
\bauthor{\bsnm{Patel}, \binits{P.}},
\bauthor{\bsnm{Jani}, \binits{A.B.}},
\bauthor{\bsnm{Curran}, \binits{W.J.}},
\bauthor{\bsnm{Liu}, \binits{T.}},
\bauthor{\bsnm{Yang}, \binits{X.}}:
\batitle{Deformable mr-cbct prostate registration using biomechanically constrained deep learning networks}.
\bjtitle{Medical physics}
\bvolume{48}(\bissue{1}),
\bfpage{253}--\blpage{263}
(\byear{2021})
\end{barticle}
\endbibitem

%%% 17
\bibitem[\protect\citeauthoryear{Wang et~al.}{2024}]{wang2024libr+}
\begin{bchapter}
\bauthor{\bsnm{Wang}, \binits{D.}},
\bauthor{\bsnm{Azadvar}, \binits{S.}},
\bauthor{\bsnm{Heiselman}, \binits{J.}},
\bauthor{\bsnm{Jiang}, \binits{X.}},
\bauthor{\bsnm{Miga}, \binits{M.}},
\bauthor{\bsnm{Wang}, \binits{L.}}:
\bctitle{Libr+: Improving intraoperative liver registration by learning the residual of biomechanics-based deformable registration}.
In: \bbtitle{International Conference on Medical Image Computing and Computer-Assisted Intervention (MICCAI)},
pp. \bfpage{359}--\blpage{368}
(\byear{2024}).
\bcomment{Springer}
\end{bchapter}
\endbibitem

%%% 18
\bibitem[\protect\citeauthoryear{Nir et~al.}{2013}]{nir2013model}
\begin{barticle}
\bauthor{\bsnm{Nir}, \binits{G.}},
\bauthor{\bsnm{Sahebjavaher}, \binits{R.S.}},
\bauthor{\bsnm{Kozlowski}, \binits{P.}},
\bauthor{\bsnm{Chang}, \binits{S.D.}},
\bauthor{\bsnm{Sinkus}, \binits{R.}},
\bauthor{\bsnm{Goldenberg}, \binits{S.L.}},
\bauthor{\bsnm{Salcudean}, \binits{S.E.}}:
\batitle{Model-based registration of ex vivo and in vivo mri of the prostate using elastography}.
\bjtitle{IEEE transactions on medical imaging}
\bvolume{32}(\bissue{6}),
\bfpage{1068}--\blpage{1080}
(\byear{2013})
\end{barticle}
\endbibitem

%%% 19
\bibitem[\protect\citeauthoryear{Segal et~al.}{2009}]{segal2009generalized}
\begin{bchapter}
\bauthor{\bsnm{Segal}, \binits{A.}},
\bauthor{\bsnm{Haehnel}, \binits{D.}},
\bauthor{\bsnm{Thrun}, \binits{S.}}:
\bctitle{Generalized-icp.}
In: \bbtitle{Robotics: Science and Systems},
vol. \bseriesno{2},
p. \bfpage{435}
(\byear{2009}).
\bcomment{Seattle, WA}
\end{bchapter}
\endbibitem

%%% 20
\bibitem[\protect\citeauthoryear{Hering et~al.}{2022}]{hering2022learn2reg}
\begin{barticle}
\bauthor{\bsnm{Hering}, \binits{A.}},
\bauthor{\bsnm{Hansen}, \binits{L.}},
\bauthor{\bsnm{Mok}, \binits{T.C.W.}}, \betal:
\batitle{Learn2reg: comprehensive multi-task medical image registration challenge, dataset and evaluation in the era of deep learning}.
\bjtitle{IEEE Transactions on Medical Imaging}
\bvolume{42}(\bissue{3}),
\bfpage{697}--\blpage{712}
(\byear{2022})
\end{barticle}
\endbibitem

%%% 21
\bibitem[\protect\citeauthoryear{Chen et~al.}{2023}]{chen2023towards}
\begin{barticle}
\bauthor{\bsnm{Chen}, \binits{W.}},
\bauthor{\bsnm{Kalia}, \binits{M.}},
\bauthor{\bsnm{Zeng}, \binits{Q.}},
\bauthor{\bsnm{Pang}, \binits{E.H.}},
\bauthor{\bsnm{Bagherinasab}, \binits{R.}},
\bauthor{\bsnm{Milner}, \binits{T.D.}},
\bauthor{\bsnm{Sabiq}, \binits{F.}},
\bauthor{\bsnm{Prisman}, \binits{E.}},
\bauthor{\bsnm{Salcudean}, \binits{S.E.}}:
\batitle{Towards transcervical ultrasound image guidance for transoral robotic surgery}.
\bjtitle{International Journal of Computer Assisted Radiology and Surgery}
\bvolume{18}(\bissue{6}),
\bfpage{1061}--\blpage{1068}
(\byear{2023})
\end{barticle}
\endbibitem

%%% 22
\bibitem[\protect\citeauthoryear{Clark et~al.}{2013}]{clark2013cancer}
\begin{barticle}
\bauthor{\bsnm{Clark}, \binits{K.}},
\bauthor{\bsnm{Vendt}, \binits{B.}},
\bauthor{\bsnm{Smith}, \binits{K.}},
\bauthor{\bsnm{Freymann}, \binits{J.}},
\bauthor{\bsnm{Kirby}, \binits{J.}},
\bauthor{\bsnm{Koppel}, \binits{P.}},
\bauthor{\bsnm{Moore}, \binits{S.}},
\bauthor{\bsnm{Phillips}, \binits{S.}},
\bauthor{\bsnm{Maffitt}, \binits{D.}},
\bauthor{\bsnm{Pringle}, \binits{M.}},
\bauthor{\bsnm{Tarbox}, \binits{L.}},
\bauthor{\bsnm{Prior}, \binits{F.}}:
\batitle{The cancer imaging archive (tcia): maintaining and operating a public information repository}.
\bjtitle{Journal of digital imaging}
\bvolume{26},
\bfpage{1045}--\blpage{1057}
(\byear{2013})
\end{barticle}
\endbibitem

%%% 23
\bibitem[\protect\citeauthoryear{Antonelli et~al.}{2022}]{simpson2019large}
\begin{barticle}
\bauthor{\bsnm{Antonelli}, \binits{M.}},
\bauthor{\bsnm{Reinke}, \binits{A.}},
\bauthor{\bsnm{Bakas}, \binits{S.}},
\bauthor{\bsnm{Farahani}, \binits{K.}},
\bauthor{\bsnm{Kopp-Schneider}, \binits{A.}},
\bauthor{\bsnm{Landman}, \binits{B.A.}},
\bauthor{\bsnm{Litjens}, \binits{G.}},
\bauthor{\bsnm{Menze}, \binits{B.}},
\bauthor{\bsnm{Ronneberger}, \binits{O.}},
\bauthor{\bsnm{Summers}, \binits{R.M.}}, \betal:
\batitle{The medical segmentation decathlon}.
\bjtitle{Nature communications}
\bvolume{13}(\bissue{1}),
\bfpage{4128}
(\byear{2022})
\end{barticle}
\endbibitem

%%% 24
\bibitem[\protect\citeauthoryear{Zeng et~al.}{2021}]{zeng2020TMI}
\begin{barticle}
\bauthor{\bsnm{Zeng}, \binits{Q.}},
\bauthor{\bsnm{Honarvar}, \binits{M.}},
\bauthor{\bsnm{Schneider}, \binits{C.}},
\bauthor{\bsnm{Mohammad}, \binits{S.K.}},
\bauthor{\bsnm{Lobo}, \binits{J.}},
\bauthor{\bsnm{Pang}, \binits{E.H.T.}},
\bauthor{\bsnm{Lau}, \binits{K.T.}},
\bauthor{\bsnm{Hu}, \binits{C.}},
\bauthor{\bsnm{Jago}, \binits{J.}},
\bauthor{\bsnm{Erb}, \binits{S.R.}},
\bauthor{\bsnm{Rohling}, \binits{R.}},
\bauthor{\bsnm{Salcudean}, \binits{S.E.}}:
\batitle{Three-dimensional multi-frequency shear wave absolute vibro-elastography (3d s-wave) with a matrix array transducer: Implementation and preliminary in vivo study of the liver}.
\bjtitle{IEEE Transactions on Medical Imaging}
\bvolume{40}(\bissue{2}),
\bfpage{648}--\blpage{660}
(\byear{2021})
\end{barticle}
\endbibitem

%%% 25
\bibitem[\protect\citeauthoryear{Zeng et~al.}{2024}]{zeng2024validation}
\begin{botherref}
\oauthor{\bsnm{Zeng}, \binits{Q.}},
\oauthor{\bsnm{Mohammad}, \binits{S.}},
\oauthor{\bsnm{Aleef}, \binits{T.A.}},
\oauthor{\bsnm{Honarvar}, \binits{M.}},
\oauthor{\bsnm{Schneider}, \binits{C.}},
\oauthor{\bsnm{Pang}, \binits{E.H.}},
\oauthor{\bsnm{Jago}, \binits{J.}},
\oauthor{\bsnm{Ramji}, \binits{A.}},
\oauthor{\bsnm{Yoshida}, \binits{E.M.}},
\oauthor{\bsnm{Rohling}, \binits{R.}},
\oauthor{\bsnm{Salcudean}, \binits{S.}}:
Validation of volumetric multi-frequency shear wave vibro-elastography with matrix array transducer for the in vivo liver.
IEEE Transactions on Ultrasonics, Ferroelectrics, and Frequency Control
(2024)
\end{botherref}
\endbibitem

%%% 26
\bibitem[\protect\citeauthoryear{Gatti and Khallaghi}{2022}]{gatti2022pycpd}
\begin{barticle}
\bauthor{\bsnm{Gatti}, \binits{A.A.}},
\bauthor{\bsnm{Khallaghi}, \binits{S.}}:
\batitle{Pycpd: Pure numpy implementation of the coherent point drift algorithm}.
\bjtitle{Journal of Open Source Software}
\bvolume{7}(\bissue{80}),
\bfpage{4681}
(\byear{2022})
\end{barticle}
\endbibitem

%%% 27
\bibitem[\protect\citeauthoryear{Wang and Solomon}{2019}]{wang2019deep}
\begin{bchapter}
\bauthor{\bsnm{Wang}, \binits{Y.}},
\bauthor{\bsnm{Solomon}, \binits{J.M.}}:
\bctitle{Deep closest point: Learning representations for point cloud registration}.
In: \bbtitle{Proceedings of the IEEE/CVF International Conference on Computer Vision (ICCV)},
pp. \bfpage{3523}--\blpage{3532}
(\byear{2019})
\end{bchapter}
\endbibitem

%%% 28
\bibitem[\protect\citeauthoryear{Evan et~al.}{2022}]{evan2022keymorph}
\begin{bchapter}
\bauthor{\bsnm{Evan}, \binits{M.Y.}},
\bauthor{\bsnm{Wang}, \binits{A.Q.}},
\bauthor{\bsnm{Dalca}, \binits{A.V.}},
\bauthor{\bsnm{Sabuncu}, \binits{M.R.}}:
\bctitle{Keymorph: Robust multi-modal affine registration via unsupervised keypoint detection}.
In: \bbtitle{Medical Imaging with Deep Learning}
(\byear{2022})
\end{bchapter}
\endbibitem

%%% 29
\bibitem[\protect\citeauthoryear{Chen et~al.}{2022}]{chen2022transmorph}
\begin{barticle}
\bauthor{\bsnm{Chen}, \binits{J.}},
\bauthor{\bsnm{Frey}, \binits{E.C.}},
\bauthor{\bsnm{He}, \binits{Y.}},
\bauthor{\bsnm{Segars}, \binits{W.P.}},
\bauthor{\bsnm{Li}, \binits{Y.}},
\bauthor{\bsnm{Du}, \binits{Y.}}:
\batitle{Transmorph: Transformer for unsupervised medical image registration}.
\bjtitle{Medical image analysis}
\bvolume{82},
\bfpage{102615}
(\byear{2022})
\end{barticle}
\endbibitem

%%% 30
\bibitem[\protect\citeauthoryear{Gustafsson and McBain}{2020}]{skfem2020}
\begin{barticle}
\bauthor{\bsnm{Gustafsson}, \binits{T.}},
\bauthor{\bsnm{McBain}, \binits{G.D.}}:
\batitle{scikit-fem: A {P}ython package for finite element assembly}.
\bjtitle{Journal of Open Source Software}
\bvolume{5}(\bissue{52}),
\bfpage{2369}
(\byear{2020})
\end{barticle}
\endbibitem

%%% 31
\bibitem[\protect\citeauthoryear{Isensee et~al.}{2021}]{isensee2021nnu}
\begin{barticle}
\bauthor{\bsnm{Isensee}, \binits{F.}},
\bauthor{\bsnm{Jaeger}, \binits{P.F.}},
\bauthor{\bsnm{Kohl}, \binits{S.A.}},
\bauthor{\bsnm{Petersen}, \binits{J.}},
\bauthor{\bsnm{Maier-Hein}, \binits{K.H.}}:
\batitle{nnu-net: a self-configuring method for deep learning-based biomedical image segmentation}.
\bjtitle{Nature methods}
\bvolume{18}(\bissue{2}),
\bfpage{203}--\blpage{211}
(\byear{2021})
\end{barticle}
\endbibitem

\end{thebibliography}



\begin{thebibliography}{1}
\ifx \bisbn   \undefined \def \bisbn  #1{ISBN #1}\fi
\ifx \binits  \undefined \def \binits#1{#1}\fi
\ifx \bauthor  \undefined \def \bauthor#1{#1}\fi
\ifx \batitle  \undefined \def \batitle#1{#1}\fi
\ifx \bjtitle  \undefined \def \bjtitle#1{#1}\fi
\ifx \bvolume  \undefined \def \bvolume#1{\textbf{#1}}\fi
\ifx \byear  \undefined \def \byear#1{#1}\fi
\ifx \bissue  \undefined \def \bissue#1{#1}\fi
\ifx \bfpage  \undefined \def \bfpage#1{#1}\fi
\ifx \blpage  \undefined \def \blpage #1{#1}\fi
\ifx \burl  \undefined \def \burl#1{\textsf{#1}}\fi
\ifx \doiurl  \undefined \def \doiurl#1{\url{https://doi.org/#1}}\fi
\ifx \betal  \undefined \def \betal{\textit{et al.}}\fi
\ifx \binstitute  \undefined \def \binstitute#1{#1}\fi
\ifx \binstitutionaled  \undefined \def \binstitutionaled#1{#1}\fi
\ifx \bctitle  \undefined \def \bctitle#1{#1}\fi
\ifx \beditor  \undefined \def \beditor#1{#1}\fi
\ifx \bpublisher  \undefined \def \bpublisher#1{#1}\fi
\ifx \bbtitle  \undefined \def \bbtitle#1{#1}\fi
\ifx \bedition  \undefined \def \bedition#1{#1}\fi
\ifx \bseriesno  \undefined \def \bseriesno#1{#1}\fi
\ifx \blocation  \undefined \def \blocation#1{#1}\fi
\ifx \bsertitle  \undefined \def \bsertitle#1{#1}\fi
\ifx \bsnm \undefined \def \bsnm#1{#1}\fi
\ifx \bsuffix \undefined \def \bsuffix#1{#1}\fi
\ifx \bparticle \undefined \def \bparticle#1{#1}\fi
\ifx \barticle \undefined \def \barticle#1{#1}\fi
\bibcommenthead
\ifx \bconfdate \undefined \def \bconfdate #1{#1}\fi
\ifx \botherref \undefined \def \botherref #1{#1}\fi
\ifx \url \undefined \def \url#1{\textsf{#1}}\fi
\ifx \bchapter \undefined \def \bchapter#1{#1}\fi
\ifx \bbook \undefined \def \bbook#1{#1}\fi
\ifx \bcomment \undefined \def \bcomment#1{#1}\fi
\ifx \oauthor \undefined \def \oauthor#1{#1}\fi
\ifx \citeauthoryear \undefined \def \citeauthoryear#1{#1}\fi
\ifx \endbibitem  \undefined \def \endbibitem {}\fi
\ifx \bconflocation  \undefined \def \bconflocation#1{#1}\fi
\ifx \arxivurl  \undefined \def \arxivurl#1{\textsf{#1}}\fi
\csname PreBibitemsHook\endcsname

%%% 1
\bibitem[\protect\citeauthoryear{Brock et~al.}{2005}]{brock2005accuracy}
\begin{barticle}
\bauthor{\bsnm{Brock}, \binits{K.}},
\bauthor{\bsnm{Sharpe}, \binits{M.}},
\bauthor{\bsnm{Dawson}, \binits{L.}},
\bauthor{\bsnm{Kim}, \binits{S.}},
\bauthor{\bsnm{Jaffray}, \binits{D.}}:
\batitle{Accuracy of finite element model-based multi-organ deformable image registration}.
\bjtitle{Medical physics}
\bvolume{32}(\bissue{6Part1}),
\bfpage{1647}--\blpage{1659}
(\byear{2005})
\end{barticle}
\endbibitem

\end{thebibliography}

\end{document}

\title[Supplementary Materials]{Supplementary Materials - SemICP: Semantic Non-Rigid Point Cloud
Registration with Elastic Energy Regularization}

\author*[1,2]{\fnm{Wanwen} \sur{Chen}}\email{wanwenc@ece.ubc.ca}

\author[1]{\fnm{Qi} \sur{Zeng}}

\author[2]{\fnm{Carson} \sur{Studders}}

\author[3]{\fnm{Zongze} \sur{Li}}

\author[2]{\fnm{Jamie J.Y.} \sur{Kwon}}

\author[3]{\fnm{Tara} \sur{Kemper}}

\author[2]{\fnm{Emily H.T.} \sur{Pang}}

\author[2]{\fnm{Eitan} \sur{Prisman}}

\author[1,3]{\fnm{Septimiu E.} \sur{Salcudean}}

\affil*[1]{\orgdiv{Department of Electrical and Computer Engineering}, \orgname{University of British Columbia}, \orgaddress{\city{Vancouver}, \state{BC}, \country{Canada}}}

\affil[2]{\orgdiv{Faculty of Medicine}, \orgname{University of British Columbia}, \orgaddress{\city{Vancouver}, \state{BC}, \country{Canada}}}

\affil[3]{\orgdiv{School of Biomedical Engineering}, \orgname{University of British Columbia}, \orgaddress{\city{Vancouver}, \state{BC}, \country{Canada}}}

\maketitle

\section{Regularization Terms Ablation Studies}
\label{sec:supp_ablation}

We first formulated SemICP with a regularization term on deformation with a Navier--Lam\'e (NL) term (Eq. 3), a gradient (GR) term (Eq.~\ref{eq:grad}), and a displacement-magnitude (MAG) term (Eq.~\ref{eq:mag}).

\begin{equation}\label{eq:grad}
Reg_{grad}(\bm{D}) = \frac{\Delta}{N_C}\sum_{\bm{c}}^{\bm{C}}||\frac{\partial \bm{D}}{\partial x_1}||_2 + ||\frac{\partial \bm{D}}{\partial x_2}||_2 + ||\frac{\partial \bm{D}}{\partial x_3}||_2 
\end{equation}
\begin{equation}\label{eq:mag}
Reg_{mag}(\bm{D})  = \frac{1}{N_C}\sum_{i}^{N_C}||\bm{d_{\bm{c}_i}}||_2 
\end{equation}

With the simulated phantom, we ran SemICP with different combinations of the regularization term and regularization weights on the 18 phantoms. 
Table~\ref{tab:supp_ablation} shows that the interior error stays close to that of no registration (1.66--1.85 vs 1.85\,mm) with $Reg_{els}+Reg_{grad}+Reg_{mag}$, and folding appears when $\alpha\leq0.1$. 
We then experimented with adding the gradient term to the Navier--Lam\'e energy, but it still doubles the interior error at the best weight (0.55 vs 0.25\,mm).

\begin{table}[h]
\caption{Phantom, regularization terms: mean $\pm$ standard deviation over 18 registrations (2 loads $\times$ 3 magnitudes $\times$ 3 samplings). Best weight $\alpha$ per combination in bold (selected with the ground truth).}
\label{tab:supp_ablation}
\centering
\setlength{\tabcolsep}{3pt}
\begin{tabular}{lccccc}
\hline
Regularizer & $\alpha$ & Interior (mm) & Surface (mm) & $|\Delta\det J|$ & Folding (\%)  \\
\hline
None (no registration)      & -- & 1.85 $\pm$ 0.99 & 1.86 $\pm$ 1.00 & 0.004 & 0.00 \\
\hline
NL (ours)                   & 0.01 & 1.00 $\pm$ 0.49 & 1.58 $\pm$ 0.75 & 0.092 & 0.05  \\
                            & 0.1 & 0.84 $\pm$ 0.46 & 1.39 $\pm$ 0.70 & 0.047 & 0.01  \\
                            & 1 & 0.60 $\pm$ 0.35 & 0.97 $\pm$ 0.46 & 0.015 & 0.00  \\
                            & \textbf{10} & \textbf{0.25 $\pm$ 0.17} & \textbf{0.35 $\pm$ 0.21} & \textbf{0.004} & \textbf{0.00} \\
                            & 100 & 0.76 $\pm$ 0.42 & 0.88 $\pm$ 0.49 & 0.004 & 0.00 \\
                            \hline
GR + NL                     & 0.01 & 0.88 $\pm$ 0.47 & 0.93 $\pm$ 0.50 & 0.051 & 0.00\\
                            & \textbf{0.1} & \textbf{0.55 $\pm$ 0.27} & \textbf{0.62 $\pm$ 0.35} & \textbf{0.011} & \textbf{0.00}  \\
                            & 1 & 1.03 $\pm$ 0.57 & 1.19 $\pm$ 0.66 & 0.005 & 0.00  \\
                            & 10 & 1.53 $\pm$ 0.84 & 1.63 $\pm$ 0.89 & 0.005 & 0.00  \\
                            & 100 & 1.76 $\pm$ 0.97 & 1.80 $\pm$ 0.99 & 0.005 & 0.00 \\
                            \hline
GR + NL + MAG               & 0.0001 & 1.75 $\pm$ 0.85 & 1.71 $\pm$ 0.76 & 0.161 & 0.79  \\
                            & 0.001 & 1.71 $\pm$ 0.86 & 1.53 $\pm$ 0.75 & 0.130 & 0.33  \\
                            & \textbf{0.01} & \textbf{1.66 $\pm$ 0.85} & \textbf{1.26 $\pm$ 0.67} & \textbf{0.098} & \textbf{0.19} \\
                            & 0.1 & 1.76 $\pm$ 0.91 & 1.57 $\pm$ 0.74 & 0.038 & 0.01  \\
                            & 1 & 1.85 $\pm$ 0.99 & 1.86 $\pm$ 1.00 & 0.005 & 0.00  \\
                            & 10 & 1.85 $\pm$ 0.99 & 1.86 $\pm$ 1.00 & 0.005 & 0.00  \\
                            & 100 & 1.85 $\pm$ 0.99 & 1.87 $\pm$ 1.00 & 0.005 & 0.00  \\
                            \hline
\end{tabular}
\end{table}

\subsection{Ablation Studies on AbdominalMRCT}
\begin{figure*}
    \centering
    \includegraphics[width=\textwidth]{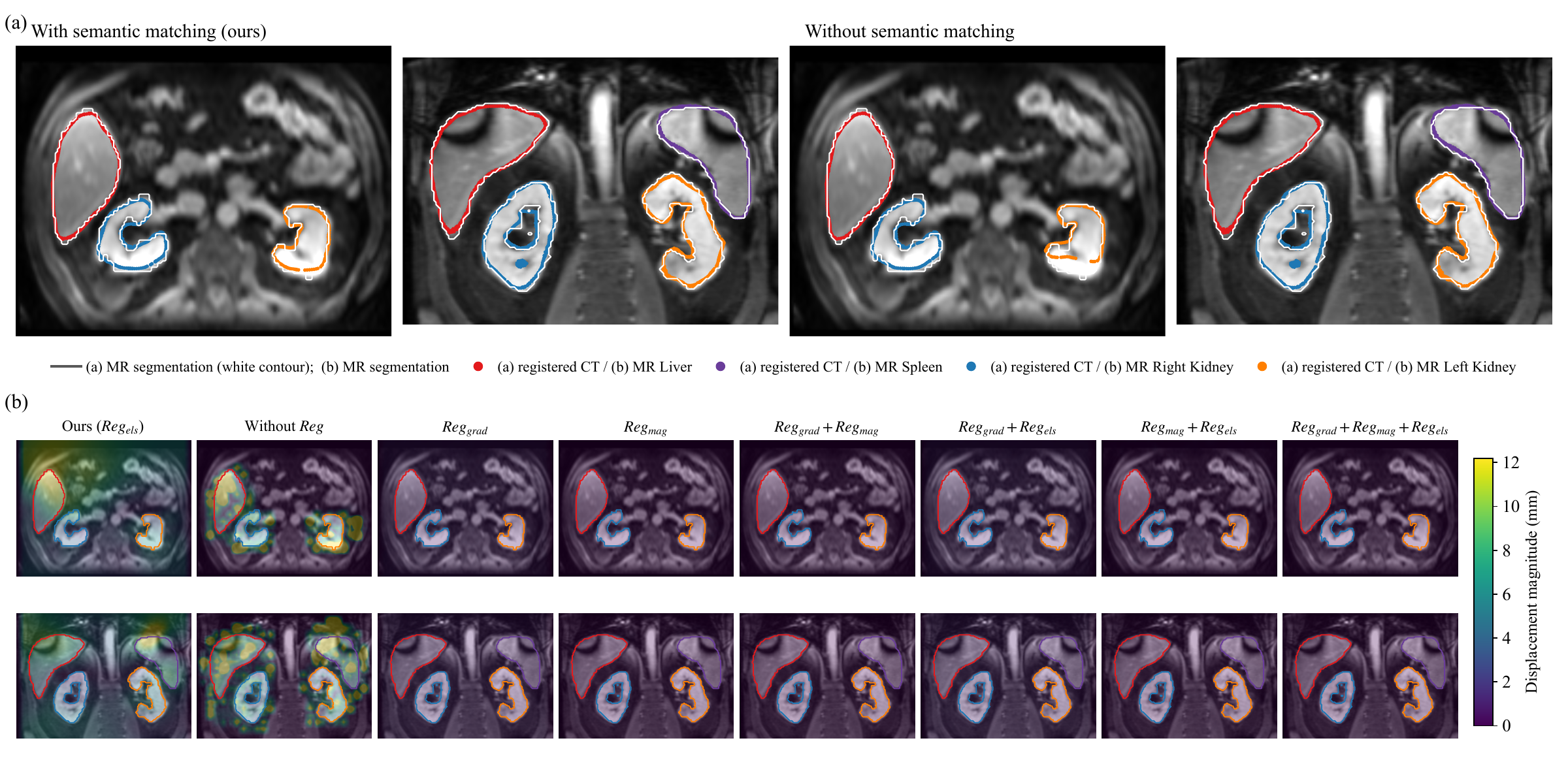}
    \caption{Ablation on AbdominalMRCT (MR axial and coronal slices through the left kidney). (a) SemICP with (left) and without (right) semantic matching. White contours: MR segmentation; colored points: registered CT surface points (red, liver; purple, spleen; blue, right kidney; orange, left kidney). 
    (b) Magnitude of the displacement field (mm, shared color scale) for each regularization configuration, overlaid on the MR slices with the MR segmentation contours. }\label{fig:ablation-semantic}
\end{figure*}

Table~\ref{tab:ablation-learn2regMRCT} reports ablation results on AbdominalMRCT, evaluating the contributions of semantic correspondence constraints and elastic energy regularization. 
Without semantic matching, but with regularization $Reg_{els}$, both HD95 and MSD degrade, demonstrating the importance of anatomically consistent correspondence estimation.
Fig.~\ref{fig:ablation-semantic}(a) shows that without semantic consistency, anatomical structure boundaries are matched less accurately.

The regularization ablation highlights the benefit of the elastic energy term $Reg_{els}$. Without regularization (Semantic w/o $Reg$), SemICP fitted the surfaces most closely, but 7.87\% of the control grid folded ($\det\bm{J}\le 0$) and SDLogJ is 2.1. With $Reg_{els}$, the control grid has no folding, though the cost was 3.7\,mm in HD95 compared with no regularization.

Among the regularized configurations, $Reg_{els}$ achieves the lowest HD95 and MSD. Other combinations suppressed the non-rigid deformation and the HD95 and MSD are high with low SDLogJ (0.0016--0.0026). We found that this is because the deformation is nearly zero, so a low SDLogJ here does not indicate a better deformation. Fig.~\ref{fig:ablation-semantic}(b) shows the corresponding displacement fields.

\begin{table*}
\centering
    \caption{HD95 and MSD (mean$\pm$std, $mm$), SDLogJ (mean$\pm$std) and folding (mean$\pm$std, \% of the grid with $J\leq0$) in ablation study on semantic matching and regularization on AbdominalMRCT ($\alpha=10$, $E=1$, $\nu=0.45$). The best is in bold, and the second best is underlined. $^*$ denotes significant difference from Semantic + $Reg_{els}$ (Ours) ($p\leq0.05$ in Wilcoxon test).}\label{tab:ablation-learn2regMRCT}
    \resizebox{\textwidth}{!}{

\begin{tabular}{ccccc}
\toprule
                                                  & HD95 & MSD & SDLogJ & Folding (\%) \\
\midrule
Initial                                           & $38.45\pm19.41$ & $17.08\pm10.39$ & N/A & N/A \\
w/o semantic matching + $Reg_{els}$                          & $11.42\pm9.03^*$ & $3.56\pm1.28^*$ & $0.015\pm0.007$ & $\mathbf{0.00\pm0.00}$ \\
Semantic  w/o $Reg$                               & $\mathbf{7.41\pm11.32^*}$ & $\mathbf{1.57\pm1.47^*}$ & $2.1\pm0.9^*$ & $7.87\pm3.02^*$ \\
Semantic + $Reg_{grad}$                           & $20.61\pm9.54^*$ & $7.78\pm2.92^*$ & $0.0017\pm0.0001^*$ & $\mathbf{0.00\pm0.00}$ \\
Semantic + $Reg_{mag}$                            & $20.80\pm9.88^*$ & $7.87\pm3.08^*$ & $\mathbf{0.0016\pm0.0002^*}$ & $\mathbf{0.00\pm0.00}$ \\
Semantic + $Reg_{els}$ (Ours)                     & \underline{$11.08\pm8.89$} & \underline{$3.47\pm1.28$} & $0.015\pm0.007$ & $\mathbf{0.00\pm0.00}$ \\
Semantic + $Reg_{grad} +   Reg_{mag}$             & $20.83\pm9.89^*$ & $7.88\pm3.09^*$ & $\mathbf{0.0016\pm0.0000^*}$ & $\mathbf{0.00\pm0.00}$ \\
Semantic + $Reg_{grad} +   Reg_{els}$             & $20.63\pm9.55^*$ & $7.78\pm2.92^*$ & $0.0017\pm0.0001^*$ & $\mathbf{0.00\pm0.00}$ \\
Semantic + $Reg_{mag} +   Reg_{els}$              & $20.81\pm9.88^*$ & $7.87\pm3.08^*$ & $0.0026\pm0.0000^*$ & $\mathbf{0.00\pm0.00}$ \\
Semantic + $Reg_{grad}+   Reg_{mag} + Reg_{els}$  & $20.83\pm9.89^*$ & $7.88\pm3.09^*$ & $\mathbf{0.0016\pm0.0000^*}$ & $\mathbf{0.00\pm0.00}$ \\
\bottomrule
\end{tabular}

}

\end{table*}

\section{Cross Validation Tuning}
We tuned the hyperparameters for optimization-based methods under the same protocol.
Each method has a grid of 10–15 configurations that contains its default setting except SemICP, which only has 3 configurations (Table~\ref{tab:supp_grids}). Solver settings such as iteration limits and convergence tolerances were kept at their published defaults. 
After every configuration was run on every case, we used cross-validation to select the parameters to report for a case. For each held-out fold, the configuration with the lowest mean surface distance over all cases of the remaining folds was chosen and applied to the held-out cases, except for Liver MR-US dataset. On the Liver dataset, because of the dataset size, the grid of every method was evaluated on the same tuning subset of two samples per subject (32 cases). For each held-out fold, the configuration was selected on the subset cases of the other two folds and then applied to all cases of the held-out fold.
The definition of folds is shown in Table~\ref{tab:supp_folds}, and no data are shared between tuning and evaluation. We report the most chosen parameters in Table~\ref{tab:supp_selected}.

A run was considered failed if it exceeded the time limit (30 min for liver dataset and 15 min for other datasets) or the results diverged (mean HD95$\geq10000$ mm or MSD$\geq1000$ mm). 
During selection, a failed run was scored as the unregistered input. In the reported results, failed cases are excluded from the means and counted separately (Table~\ref{tab:supp_failures}). Cases on which AMM-NRR or Fast-RNRR did not finish the initialization during tuning were not re-run in the final timed runs and are counted as failed.

Fixed parameters: ICP runs 50 iterations and starts by matching the centroids as in the default VTK implementation. FR-ICP uses its robust point-to-point variant. NR-ICP uses 20 stiffness steps, decreasing linearly from $\alpha_{\mathrm{start}}$ to $\alpha_{\mathrm{end}}$, and is initialized with ICP. AMM-NRR and Fast-RNRR use a node-sampling radius of 5 times the average edge length. SemICP uses Young's modulus $E=1$, and the elastic weight is fixed to 10 based on the results in the FEM phantoms, where the ground-truth deformation field was available. GO-ICP and the SemICP rigid stage estimate the transform from a fixed number of 1000 points per cloud, sampled evenly across labels, on AbdominalMR-CT, TORS, and HippocampusMR, and all points on Liver MR–3DUS. AMM-NRR and Fast-RNRR were not evaluated on Liver MR--3DUS because, in 8 randomly sampled tested cases from 8 subjects, their node-sampling initialization did not complete within the 15-minute limit.

\begin{table*}
\centering
\caption{Parameter search grids for the cross validation. The default is shown in bold.}
\label{tab:supp_grids}
\small
\resizebox{\textwidth}{!}{
\begin{tabular}{lllc}
\toprule
Method & Parameter & Values & Configs \\
\midrule
\multirow{2}{*}{ICP (rigid)} & number of sampled points & 	\textbf{1000} (Liver: 	\textbf{all}), 5000, all & \multirow{2}{*}{15} \\
 & landmarks matched per iteration & 	\textbf{200}, 500, 1000, 2000, 5000 & \\
\midrule
\multirow{2}{*}{CPD (rigid)} & outlier weight $w$
& \textbf{0}, 0.05, 0.1, 0.2, 0.3 & \multirow{2}{*}{15} \\
 & number of sampled points & \textbf{1000}, 3000, 5000 & \\
\midrule
\multirow{2}{*}{GO-ICP} & convergence threshold (MSE) & 0.005, \textbf{0.01} & \multirow{2}{*}{10} \\
 & trimming fraction & \textbf{0}, 0.05, 0.1, 0.2, 0.3 & \\
\midrule
\multirow{2}{*}{FR-ICP} & initial Welsch kernel width $\nu_{\mathrm{begin}}$ & 1, \textbf{3}, 10 & \multirow{2}{*}{15} \\
 & final Welsch kernel width $\nu_{\mathrm{end}}$ & 0.05, 0.1, \textbf{0.192}, 0.4, 0.8 & \\
\midrule
\multirow{2}{*}{NR-ICP} & stiffness at the start of the schedule $\alpha_{\mathrm{start}}$ & $2{\times}10^{2}$, $2{\times}10^{3}$, $\boldsymbol{2{\times}10^{4}}$, $2{\times}10^{5}$, $2{\times}10^{6}$ & \multirow{2}{*}{15} \\
 & stiffness at the end of the schedule $\alpha_{\mathrm{end}}$ & 0.1, \textbf{1}, 10 & \\
\midrule
\multirow{2}{*}{AMM-NRR} & smoothness weight $\alpha$ & 1, 10, \textbf{100}, 1000, $10^{4}$ & \multirow{2}{*}{15} \\
 & rigidity weight $\beta$ & 1, \textbf{10}, 100 & \\
\midrule
\multirow{2}{*}{Fast-RNRR} & smoothness weight $\alpha$ & 1, 10, \textbf{100}, 1000, $10^{4}$ & \multirow{2}{*}{15} \\
 & rigidity weight $\beta$ & 10, \textbf{100}, 1000 & \\
\midrule
\multirow{2}{*}{SemICP} & Poisson ratio $\nu$ & 0.45, \textbf{0.49}, 0.499 & \multirow{2}{*}{3} \\
 & regularization weight & 10 (fixed from the FEM phantom) & \\
\bottomrule
\end{tabular}
}
\end{table*}

\begin{table*}
\centering
\caption{Cross-validation fold definition and number of folds in each dataset.}
\label{tab:supp_folds}
\small
\resizebox{\textwidth}{!}{
\begin{tabular}{lccp{7.5cm}}
\toprule
Dataset & Cases & Folds & Grouping \\
\midrule
AbdominalMR-CT & 8 & 8 & leave-one-patient-out (8 patients) \\
TORS & 24 & 12 & leave-one-patient-out (24 sequences from 12 patients) \\
HippocampusMR & 60 & 30 & leave-one-pair-out: each image pair is registered in both directions, and both directions are held out together \\
Liver MR--3DUS & 160 & 3 & the three patient-level folds used to train KeyMorph and TransMorph (16 subjects $\times$ 10 samples) \\
\bottomrule
\end{tabular}
}
\end{table*}

\begin{table*}
\centering
\caption{The configuration chosen in most folds in each dataset (number of folds in parentheses). }
\label{tab:supp_selected}
\footnotesize
\setlength{\tabcolsep}{4pt}
\resizebox{\textwidth}{!}{
\begin{tabular}{llcccc}
\toprule
Method & Parameters & MR-CT (8) & TORS (12) & Hippocampus (30) & Liver (3) \\
\midrule
ICP & points, landmarks & 5000, 2000 (4) & 5000, 500 (11) & 1000, 1000 (30) & \makecell{5000, 2000; 1000, 2000;\\5000, 5000 (1 each)} \\
CPD & $w$, points & 0.3, 1000 (6) & 0.3, 1000 (12) & 0.2, 3000 (26) & 0.3, 1000 (2) \\
GO-ICP & MSE, trim & 0.005, 0.05 (4) & 0.005, 0 (11) & 0.005, 0 (29) & \makecell{0.005, 0 / 0.1 / 0.3\\(1 each)} \\
FR-ICP & $\nu_{\mathrm{begin}}$, $\nu_{\mathrm{end}}$ & 1, 0.4 (5) & 3, 0.8 (6) & 3, 0.8 (30) & 1, 0.4 (2) \\
NR-ICP & $\alpha_{\mathrm{start}} \rightarrow \alpha_{\mathrm{end}}$ & $2{\times}10^{2} \rightarrow 1$ (6) & $2{\times}10^{5} \rightarrow 1$ (11) & $2{\times}10^{6} \rightarrow 0.1$ (30) & $2{\times}10^{3} \rightarrow 0.1$ (2) \\
AMM-NRR & $\alpha$, $\beta$ & 100, 1 (6) & 100, 1 (12) & 1, 10 (30) & -- \\
Fast-RNRR & $\alpha$, $\beta$ & 100, 1000 (7) & 1000, 100 (11) & 10, 100 (28) & -- \\
SemICP & $\nu$ & 0.45 (8) & 0.499 (12) & 0.45 (30) & 0.45 (3) \\
\bottomrule
\end{tabular}
}
\end{table*}

\begin{table*}
\centering
\caption{Failed registrations in the final (tuned) runs. All other method--dataset combinations, including SemICP on all datasets, completed every case successfully.}
\label{tab:supp_failures}
\small
\resizebox{\textwidth}{!}{
\begin{tabular}{llccp{6cm}}
\toprule
Dataset & Method & Cases & Successful & Failed (reason) \\
\midrule
\multirow{2}{*}{AbdominalMR-CT} & AMM-NRR & 8 & 7 & 1 (initialization did not finish) \\
 & Fast-RNRR & 8 & 7 & 1 (initialization did not finish) \\
\midrule
\multirow{3}{*}{TORS} & AMM-NRR & 24 & 7 & 17 (initialization did not finish) \\
 & Fast-RNRR & 24 & 6 & 18 (17 initializations did not finish, 1 diverged) \\
 & NR-ICP & 24 & 23 & 1 (diverged) \\
\midrule
\multirow{2}{*}{HippocampusMR} & AMM-NRR & 60 & 58 & 2 (initialization did not finish) \\
 & Fast-RNRR & 60 & 57 & 3 (2 initialization did not finish, 1 crashed) \\
\bottomrule
\end{tabular}
}
\end{table*}

\section{Implementation Details}

The methods were run on a 12GB Nvidia GeForce RTX 3060 GPU, and were implemented in Python 3.9.18, PyTorch 2.1.2, and CUDA-11.8. 
The rigid alignment used a learning rate of 0.001, and the maximum number of iterations is 500 (100 on liver). The non-rigid refinement used a learning rate of 0.01, and the maximum number of iterations is 1000 (100 on Liver). 
The optimization was stopped early if the loss no longer improved. 
Each point's coordinate was normalized to $[-1,1]^3$, and $25\times25\times25$ control points were uniformly sampled within this normalized space.

\section{Sensitivity Study}\label{sec1}

\subsection{Sensitivity on Elasticity Values}
To evaluate the sensitivity of SemICP to the Young's modulus $E$ and Poisson's ratio $\nu$, we performed SemICP on the Liver MR-US and AbdominalMRCT datasets with varying elastic parameters, and kept other settings as in the main experiments ($\alpha=10$).
Typical $E$ for soft tissues and abdominal organs range from 1kPa to 100 kPa, with the stomach reaching 500kPa, while $\nu$ typically ranges from 0.400 to 0.499~\cite{brock2005accuracy}.
For the $E$ sensitivity analysis, we fixed $\nu=0.45$, the value selected by cross-validation, and varied $E$ across \{1kPa, 10kPa, 100kPa, 500kPa, 1MPa\}.
To assess the influence of $\nu$, we fixed $E$ at 1kPa and varied $\nu$ in \{0.35, 0.40, 0.45, 0.499\}. 

The sensitivity to Young's modulus  $E$ is presented in Table~\ref{tab:liver-sensitivity-strain} and Table~\ref{tab:sensitivity-strain}. 
As $E$ increases, the surface matching errors HD95 and MSD increase, and TRE in liver data also increases, while the spatial distortion SDLogJ decreases.
The results suggest that as tissue stiffness increases, the estimated deformation becomes more regularized, with a reduction in its ability to fit boundary conditions and the internal deformation, as measured by the TRE.

The sensitivity to Poisson's Ratio $\nu$ is shown in Table~\ref{tab:liver-sensitivity-compressibility} and Table~\ref{tab:sensitivity-compressibility}. As $\nu$ decreases, SDLogJ increases, indicating that assuming greater tissue compressibility weakens the linear elastic energy regularization, allowing larger but less regularized deformations to fit the boundary conditions. 
On AbdominalMRCT, a lower $\nu$ reduces the surface errors. However, on Liver MR-US, the TRE is the same for $\nu\leq0.45$ but the MSD reduces when $\nu$ decreases, so the closer surface fit does not translate into a more accurate interior deformation. The nearly
incompressible $\nu=0.499$ increases all errors on both datasets.

\begin{table*}
\centering
\caption{The average TRE, HD95, MSD, and SDLogJ on Liver MR-US for different Young's Modulus $E$, while $\nu=0.45$, $\alpha=10$. TRE, HD95 and MSD are reported in $mm$. $^*$ means the results are significantly different compared with $E$=1kPa under $p\leq0.05$ in a Wilcoxon test.
}\label{tab:liver-sensitivity-strain}
\begin{tabular}{cccccc}
\toprule
$E$ & 1kPa & 10kPa & 100kPa & 500kPa & 1MPa \\
\midrule
TRE & 5.51 & 6.02$^*$ & 6.27$^*$ & 6.28$^*$ & 6.28$^*$ \\
HD95 & 15.59 & 16.37$^*$ & 17.02$^*$ & 17.07$^*$ & 17.08$^*$ \\
MSD & 3.99 & 4.62$^*$ & 5.02$^*$ & 5.05$^*$ & 5.05$^*$ \\
SDLogJ & 0.013 & 0.0022$^*$ & 0.00031$^*$ & 0.00018$^*$ & 0.00017$^*$ \\
\bottomrule
\end{tabular}
\end{table*}

\begin{table*}
\centering
\caption{The average HD95, MSD and SDLogJ on AbdominalMRCT for different Young's Modulus $E$, while $\nu=0.45$, $\alpha=10$. HD95 and MSD are reported in $mm$. $^*$ means the results are significantly different compared with $E$=1kPa under $p\leq0.05$ in a Wilcoxon test.}\label{tab:sensitivity-strain}
\begin{tabular}{cccccc}
\toprule
$E$ & 1kPa & 10kPa & 100kPa & 500kPa & 1MPa \\
\midrule
HD95 & 11.08 & 16.83$^*$ & 20.11$^*$ & 20.66$^*$ & 20.75$^*$ \\
MSD & 3.47 & 6.24$^*$ & 7.59$^*$ & 7.81$^*$ & 7.85$^*$ \\
SDLogJ & 0.015 & 0.0034$^*$ & 0.00049$^*$ & 0.00023$^*$ & 0.00023$^*$ \\
\bottomrule
\end{tabular}
\end{table*}

\begin{table*}
\centering
\caption{The average TRE, HD95, MSD and SDLogJ on Liver MR-US for different Poisson's ratio $\nu$, with $E$=1kPa, $\alpha=10$. TRE, HD95 and MSD are reported in $mm$. $^*$ means the results are significantly different compared with $\nu$=0.45 under $p\leq0.05$ in a Wilcoxon test.}\label{tab:liver-sensitivity-compressibility}
\begin{tabular}{ccccc}
\toprule
$\nu$ & 0.35 & 0.40 & 0.45 & 0.499 \\
\midrule
TRE & 5.50 & 5.50 & 5.51 & 5.95$^*$ \\
HD95 & 15.61 & 15.58 & 15.59 & 16.16$^*$ \\
MSD & 3.96$^*$ & 3.97$^*$ & 3.99 & 4.42$^*$ \\
SDLogJ & 0.016$^*$ & 0.014$^*$ & 0.013 & 0.010$^*$ \\
\bottomrule
\end{tabular}
\end{table*}

\begin{table*}
\centering
\caption{The average HD95, MSD and SDLogJ on AbdominalMRCT for different Poisson's ratio $\nu$, with $E$=1kPa, $\alpha=10$. HD95 and MSD are reported in $mm$. $^*$ means the results are significantly different compared with $\nu=0.45$ under $p\leq0.05$ in a Wilcoxon test.}\label{tab:sensitivity-compressibility}
\begin{tabular}{ccccc}
\toprule
$\nu$ & 0.35 & 0.40 & 0.45 & 0.499 \\
\midrule
HD95 & 9.34$^*$ & 10.11$^*$ & 11.08 & 13.39$^*$ \\
MSD & 2.91$^*$ & 3.12$^*$ & 3.47 & 4.45$^*$ \\
SDLogJ & 0.018$^*$ & 0.017 & 0.015 & 0.014$^*$ \\
\bottomrule
\end{tabular}
\end{table*}

\subsection{Sensitivity on Misalignment and Visualizable Ratio}
To assess the sensitivity of SemICP to initial misalignments and partial visualization, we simulated cropped point clouds with various rigid misalignments on the AbdominalMRCT dataset. 
We cropped the original point clouds to retain 5\%, 10\%, 25\% and 50\% of points and added Gaussian noise. 
Different levels of rigid misalignment, with Euler rotation angles within $\pm5^\circ$, $\pm15^\circ$, and $\pm30^\circ$ were introduced. 
For each configuration, 10 different simulated partial point clouds were generated per patient, yielding 80 synthetic point cloud pairs. 
We compare SemICP-rigid method with ICP and GO-ICP to show how the semantic matching improves ICP robustness. FR-ICP was excluded due to its higher HD95 in Table 3 of the main paper, indicating less favorable performance. 
The point clouds were not downsampled before registration.

As shown in Table~\ref{tab:sensitivity-rigid}, SemICP-rigid achieves lower unidirectional HD95, MSD, and TRE compared to ICP and GO-ICP under challenging visibility conditions (5\%, 10\%, and 25\%). 
When evaluating sensitivity to initial misalignment, both ICP and GO-ICP exhibit increased registration error at rotation angles of $\pm30^\circ$, indicating sensitivity to initialization. In contrast, SemICP-rigid demonstrates improved robustness, with smaller error increases under the same conditions.
These results highlight that incorporating semantic labels enhances the accuracy and stability of the rigid registration step under limited visibility and challenging initial alignment.

\begin{table*}
\centering
\caption{The average unidirectional HD95, unidirectional MSD and TRE of all labels for different rigid ICP methods with different visible ratios of the point cloud. The numbers are reported in $mm$. The best values are bold and the second best is underlined. $^*$ means the results are significantly different compared with the best or the second-best experiments under $p\leq0.05$ in a two-sample t-test.}\label{tab:sensitivity-rigid}
\resizebox{\textwidth}{!}{

\begin{tabular}{cc|ccc|ccc|ccc|ccc}
\toprule
                      &                 & \multicolumn{3}{c|}{Init} & \multicolumn{3}{c|}{ICP} & \multicolumn{3}{c|}{GO-ICP}                      & \multicolumn{3}{c}{SemICP-rigid}                \\
Vis                   & Rot             & HD95   & MSD    & TRE    & HD95   & MSD    & TRE   & HD95           & MSD           & TRE            & HD95           & MSD           & TRE            \\
\midrule
\multirow{3}{*}{5\%}  & $\pm5^{\circ}$  & 13.60  & 6.31   & 14.17  & 43.31  & 29.53  & 59.28 & \underline{9.39}           & \underline{4.02}          & \underline{8.58}           & \textbf{9.24}$^*$   & \textbf{3.94}  & \textbf{8.23}   \\
                      & $\pm15^{\circ}$ & 32.95  & 19.52  & 42.73  & 36.20  & 23.70  & 51.99 & \underline{12.09}          & \underline{5.85}          & \underline{14.20}          & \textbf{10.12}$^*$  & \textbf{4.34}$^*$  & \textbf{10.42}$^*$  \\
                      & $\pm30^{\circ}$ & 61.06  & 43.25  & 80.26  & 42.35  & 28.89  & 61.36 & \underline{27.97}          & \underline{18.15}         & \underline{40.61}          & \textbf{13.82}$^*$  & \textbf{6.27}$^*$  & \textbf{19.75}$^*$  \\
                      \midrule
\multirow{3}{*}{10\%} & $\pm5^{\circ}$  & 13.74  & 6.22   & 13.19  & 42.49  & 28.24  & 55.73 & \underline{9.27}           & \underline{3.96}          & \underline{8.35}           & \textbf{9.23}  & \textbf{3.95} & \textbf{8.20}  \\
                      & $\pm15^{\circ}$ & 30.90  & 17.45  & 38.22  & 35.83  & 23.23  & 47.19 & \underline{11.34}          & \underline{5.20}          & \underline{12.32}          & \textbf{10.19} & \textbf{4.41} & \textbf{10.10} \\
                      & $\pm30^{\circ}$ & 55.44  & 35.60  & 70.28  & 36.21  & 22.75  & 50.35 & \underline{23.59}          & \underline{13.28}         & \underline{32.00}          & \textbf{14.55}$^*$   & \textbf{6.44}$^*$   & \textbf{18.55}$^*$   \\
                      \midrule
\multirow{3}{*}{25\%} & $\pm5^{\circ}$  & 13.73  & 6.09   & 14.11  & 17.62  & 9.49   & 20.22 & \underline{9.38}           & \underline{3.98}          & \underline{8.30}           & \textbf{9.37}$^*$   & \textbf{3.97}$^*$  & \textbf{8.22}$^*$   \\
                      & $\pm15^{\circ}$ & 33.27  & 17.21  & 40.83  & 18.84  & 9.58   & 22.65 & \underline{10.53}          & \underline{4.48}          & \underline{10.16}          & \textbf{10.43} & \textbf{4.42} & \textbf{9.90}  \\
                      & $\pm30^{\circ}$ & 65.49  & 41.65  & 78.72  & 18.43  & \underline{8.52}   & 23.26 & \underline{16.81}          & 8.88          & \underline{19.40}          & \textbf{15.41} & \textbf{6.74}$^*$  & \textbf{17.79} \\
                      \midrule
\multirow{3}{*}{50\%} & $\pm5^{\circ}$  & 14.92  & 6.49   & 14.99  & 9.77   & 4.13   & 8.62  & \underline{9.55}           & \underline{4.03}          & \underline{8.34}           & \textbf{9.54}  & \textbf{4.03} & \textbf{8.31}$^*$   \\
                      & $\pm15^{\circ}$ & 33.74  & 16.19  & 38.32  & 11.16  & 4.70   & 10.26 & \textbf{10.82} & \textbf{4.54} & \textbf{9.78}  & \underline{10.91}$^*$        & \underline{4.58}$^*$          & \underline{9.88}           \\
                      & $\pm30^{\circ}$ & 72.80  & 43.07  & 81.74  & \underline{14.02}  & \underline{5.84}   & \underline{13.89} & \textbf{13.16} & \textbf{5.42} & \textbf{12.29} & 16.70$^*$           & 7.10$^*$          & 18.05$^*$         \\
                      \bottomrule
\end{tabular}
}
\end{table*}

\bibliography{sn-bibliography-supplemental}% common bib file